\documentclass[11pt]{article}

\RequirePackage[english]{babel}
\RequirePackage{amsthm}
\RequirePackage{amsmath}
\RequirePackage{graphicx}
\RequirePackage[countmax]{subfloat}
\RequirePackage{subfig}
\RequirePackage{natbib}
\RequirePackage{setspace}
\RequirePackage{geometry} % use with 11pt
\RequirePackage{tabularx}
\RequirePackage{booktabs}
\RequirePackage{mathtools}
\mathtoolsset{mathic=true,showonlyrefs} 
\RequirePackage{tikz}
\RequirePackage{enumitem}
\usepackage{setspace}
\usepackage{multirow}
\usepackage{adjustbox}
\usepackage{lscape}
\usepackage{booktabs}
\usepackage{threeparttable}
\usepackage{adjustbox}
\usepackage{pdflscape}
\usepackage{array}
\usepackage{siunitx}
\usepackage[toc]{appendix} % Handles appendices and their TOC inclusion

\RequirePackage[sf,bf]{titlesec}
\RequirePackage{fancyhdr}
\RequirePackage{caption}
\RequirePackage[original]{abstract}

\RequirePackage{microtype}

\newtheorem{definition}{Definition}
\newtheorem{theorem}{Theorem}
\newtheorem{assumption}{Assumption}
\newtheorem{proposition}{Proposition}

\newtheorem{remark}{Remark}

\allowdisplaybreaks[4]

\newcommand{\E}{\mathbb{E}}

\numberwithin{equation}{section}

\RequirePackage{amssymb}
\RequirePackage{mlmodern}
\RequirePackage[scr=boondoxo, bb=pazo, frak=boondox]{mathalpha}
\RequirePackage{bm}

\makeatletter
\newcommand*{\transp}{
	{\mathpalette\@transp{}}
}
\newcommand*{\@transp}[2]{
	\raisebox{\depth}{$\m@th#1\intercal$}
}
\makeatother

\RequirePackage{hyperref}
\hypersetup{%
	colorlinks = true,
	linkcolor  = blue,
	citecolor  = blue,
}

\usepackage{ifluatex}

\newcommand{\F}{\mathcal{F}}

\newcommand{\X}{\mathcal{X}}

\DeclareMathOperator*{\argmin}{arg\,min}

\usepackage{stackengine}

\title{\textbf{\textsf{Scale-Insensitive Neural Network Significance Tests}}
}

\author{Hasan Fallahgoul\thanks{Hasan Fallahgoul, Monash University, School of Mathematics and Centre for Quantitative Finance and Investment Strategies, 9 Rainforest Walk, 3800 Victoria, Australia. E-mail: \texttt{hasan.fallahgoul@monash.edu}.} \\ {\small Monash University} 
}
	
\date{\today }

\begin{document}

\maketitle

\thispagestyle{empty}

\begin{abstract}
This paper develops a scale-insensitive framework for neural network significance testing, substantially generalizing existing approaches through three key innovations. First, we replace metric entropy calculations with Rademacher complexity bounds, enabling the analysis of neural networks without requiring bounded weights or specific architectural constraints. Second, we weaken the regularity conditions on the target function to require only Sobolev space membership $H^s([-1,1]^d)$ with $s > d/2$, significantly relaxing previous smoothness assumptions while maintaining optimal approximation rates. Third, we introduce a modified sieve space construction based on moment bounds rather than weight constraints, providing a more natural theoretical framework for modern deep learning practices. Our approach achieves these generalizations while preserving optimal convergence rates and establishing valid asymptotic distributions for test statistics. The technical foundation combines localization theory, sharp concentration inequalities, and scale-insensitive complexity measures to handle unbounded weights and general Lipschitz activation functions. This framework better aligns theoretical guarantees with contemporary deep learning practice while maintaining mathematical rigor. 
% The proofs employ modern empirical process theory, using Rademacher complexity for both upper bounds and asymptotic analysis, providing a more direct route to establishing the asymptotic properties of test statistics while accommodating flexible neural network architectures.

\end{abstract}


% \tableofcontents

\newpage

\setstretch{1}

\section{Introduction}

Deep neural networks have emerged as powerful tools for nonparametric estimation, consistently demonstrating superior predictive performance across many domains \citep{goodfellow2016deep,anthony1999neural}. However, their effectiveness comes at the cost of interpretability - while these models can capture complex nonlinear relationships, understanding which input variables significantly influence their predictions remains challenging. Recent work has begun to address these challenges through both scale-insensitive analysis and variable significance testing.

A significant methodological advance came from \cite{farrell2021deep}, who established a scale-insensitive framework for analyzing neural networks. Their approach avoids restrictive weight constraints through careful localization arguments, enabling rigorous statistical analysis of modern deep learning practices. In parallel, \cite{horel2020significance} developed the first theoretically-based statistical test for assessing variable impact in regression using a one-layer sigmoid network. \cite{fallahgoul2024asset} extended \cite{horel2020significance}'s work to multi-layer perceptrons (MLPs), which are generally preferred over one-layer networks due to their superior approximative power. However, their framework, based on metric entropy analysis, still requires several restrictive assumptions including bounded weights and specific architectural constraints.

This paper bridges these two lines of research by extending the framework of \cite{fallahgoul2024asset} using scale-insensitive measures inspired by \cite{farrell2021deep}. Specifically, we replace metric entropy calculations with Rademacher complexity bounds, drawing on results from \cite{KoltchinskiiPancheko2000} and \cite{Bartlett2005}. Our methodology closely aligns with \cite{farrell2021deep}, but addresses a different challenge: while they focus on causal inference through partial effects, we develop rigorous significance tests for neural network inputs.

Our theoretical framework rests on three key innovations. First, following \cite{farrell2021deep}, we use scale-insensitive complexity measures that enable the analysis of neural networks without requiring bounded weights or specific architectural constraints. Second, we weaken the regularity conditions on the target function to require only Sobolev space membership $H^s([-1,1]^d)$ with $s > d/2$, substantially relaxing the conditions in \cite{fallahgoul2024asset}. Third, we introduce a modified sieve space construction based on moment bounds rather than weight constraints, providing a more natural theoretical framework for contemporary deep learning practices.

Our analysis proceeds through careful localization arguments similar to those in \cite{farrell2021deep}. We first establish consistency of the neural network estimator and derive non-asymptotic bounds on its convergence rate. Using tools from empirical process theory \citep{vaart1996weak}, we then show that the test statistic has a well-defined limiting distribution under the null hypothesis. The key technical challenge lies in controlling the complexity of the function class without imposing artificial constraints on the network architecture, which we address through Rademacher complexity bounds.

Our theoretical results build upon and complement existing approaches in deep learning theory. In particular, our work and \cite{farrell2021deep} both achieve optimal rates through distinct technical approaches that illuminate different aspects of neural network estimation. \cite{farrell2021deep} develop a precise architecture-specific analysis, building on \cite{yarotsky2017error,yarotsky2018optimal}'s explicit ReLU constructions to provide sharp bounds and detailed insights into network expressivity. Their careful control of network parameters yields tight constants and explicit optimization guidance, revealing intricate connections between architecture and statistical complexity.

Our framework takes a complementary path, employing general functional analytic methods and scale-insensitive measures through Rademacher complexity. This approach accommodates broader network architectures and activation functions under moment conditions, trading some constant-specificity for increased generality. Where \cite{farrell2021deep} leverage explicit ReLU constructions for precise parameter efficiency bounds, we use general Sobolev space approximation properties to establish optimal rates across a broader function class. This contrast reflects a fundamental tension in deep learning theory between architecture-specific precision and architecture-agnostic flexibility.

Both approaches utilize localization techniques to different effect. \cite{farrell2021deep}'s detailed shell-by-shell analysis provides sharp control of the approximation-estimation trade-off through careful handling of unbounded weights. Our more streamlined approach focuses on concentration inequalities, yielding simpler proofs while maintaining optimal rates. These complementary perspectives—\cite{farrell2021deep}'s architectural precision and our functional generality—contribute to a deeper understanding of the relationship between network structure and statistical estimation.

This paper contributes to several strands of literature. First, we advance the theoretical understanding of deep neural networks by providing rigorous statistical inference tools with weaker assumptions than previous work \citep{horel2020significance, fallahgoul2024asset}. Second, we extend the scale-insensitive analysis of \cite{farrell2021deep} to the problem of variable significance testing. Third, we contribute to the broader literature on significance testing in nonparametric models by developing methods applicable to modern machine learning architectures.

% Our approach to neural network analysis differs markedly from existing work. Unlike \cite{fallahgoul2024asset}, we do not restrict their class of network architectures to have bounded weights for each node, which allows for a richer set of approximating possibilities. Instead, following \cite{farrell2021deep}, we use a localized analysis that derives bounds based on scale-insensitive complexity measures. However, unlike \cite{farrell2021deep} who focus on establishing nonasymptotic bounds for causal parameters, our work develops asymptotic theory for significance testing.

% The remainder of the paper is organized as follows. Section \ref{sec: Setting} establishes the regression framework and presents the neural network architectures we study. Section \ref{sec: Inference for Scale-Insensitive Neural Network Tests} develops the main theoretical results, covering inferences based on these MLPs. Specifically, Subsection \ref{subsection: Consistency} discusses the probabilistic convergence of smooth MLPs and the estimation rate of the regression target function. This convergence rate serves as a basis for constructing the statistical test in Subsection \ref{subsection: Asymptotic Distribution}, which is presented along with its limiting distribution. Subsection \ref{subsection: Computing Distribution} provides a discretization approach to compute the asymptotic distribution of the test statistic. All proofs are relegated to the appendix.

The remainder of the paper is organized as follows. Section \ref{sec: Setting} establishes the foundational elements of our scale-insensitive framework, including the regression setting, modified regularity conditions, hypothesis testing framework, neural network architecture, Rademacher complexity measures, and estimation methodology. Section \ref{sec: Inference for Scale-Insensitive Neural Network Tests} develops the main theoretical results for scale-insensitive neural network tests. Specifically, Subsection \ref{subsection: Consistency of Scale-Insensitive MLPs} establishes consistency results for neural networks under our relaxed conditions, Subsection \ref{subsection: Scale-Insensitive Sieve Spaces} introduces scale-insensitive sieve spaces that replace traditional metric entropy calculations, and Subsection \ref{subsection: Asymptotic Distribution of Test Statistics} presents the asymptotic distribution theory for our test statistics. Section \ref{sec_conclusion} provides the conclusions of this study. All proofs are relegated to the appendix.

\section{Scale-Insensitive Framework and Preliminaries}

In this section, we establish the foundational elements of our scale-insensitive framework for neural network significance testing. We begin by introducing the regression framework and necessary assumptions, followed by the hypothesis testing framework and the architecture of deep neural networks under consideration. Readers who are interested in the proofs may refer to appendixes.

\subsection{Regression Framework and Scale-Insensitive Setting}\label{sec: Setting}

Let $(\Omega, \F, P)$ be a probability space where we observe the relationship between a response variable $Y$ and covariates $X \in \X := [-1,1]^d$ through the regression model
\begin{equation}
Y = f_\star(X) + \varepsilon.
\end{equation}

\begin{assumption}[Data Generation]\label{eq:Data Generation}
We assume that:
\begin{enumerate}[label=\alph*.]
    \item The response variable $Y$ has absolute value bounded above by $M_Y > 0$
    \item The covariate vector $X$ is continuously distributed according to $P$
    \item The error term $\varepsilon$ is independent of $X$ with $\E[\varepsilon] = 0$ and $\E[\varepsilon^2] = \sigma^2 < \infty$
    \item The joint law of $(X,\varepsilon)$ is absolutely continuous with respect to Lebesgue measure $\Delta$.
\end{enumerate}
\end{assumption}

Unlike previous frameworks, \cite{fallahgoul2024asset, horel2020significance} that rely on bounded weights and strict architectural constraints, we introduce scale-insensitive conditions through moment bounds. For the target function $f_\star$, we require

\begin{equation}
\E[f_\star(X)^2] \leq M \text{ and } \|f_\star\|_\infty + \sum_{\|\alpha\|_1=1} \|\nabla^\alpha f_\star\|_\infty \leq 2M
\end{equation}
where $M$ is a positive constant and $\alpha$ represents multi-index notation for derivatives.

\subsection{Modified Regularity Conditions}

We now introduce weaker regularity conditions that replace the classical smoothness and structural assumptions in \cite{fallahgoul2024asset, horel2020significance}.

\begin{assumption}[Modified Regularity]\label{assumption: Modified Regularity}
The target function $f_\star$ satisfies
\begin{enumerate}[label=\alph*.]
    \item $f_\star \in H^s([-1,1]^d)$ for $s > d/2$
    \item $\|f_\star\|_{H^s}^2 = \sum_{|\alpha| \leq s} \|\partial^\alpha f\|_{L^2}^2 < \infty$
\end{enumerate}
where $H^s([-1,1]^d)$ denotes the Sobolev space of order $s$.\footnote{Discussion of Hölder, Sobolev, and Besov spaces can be found in \cite{gine2021mathematical}.}
\end{assumption}

Our framework substantially modifies the original structural assumptions from \cite{fallahgoul2024asset}, building on the scale-insensitive approach pioneered by \cite{farrell2021deep}. In particular, let us examine the original Assumptions 2.2 and 2.3, which require:
\begin{enumerate}[label=\alph*.]
    \item The target function $f_\star$ to belong to $\mathcal{C}^{\lfloor d/2\rfloor + 2}([-1,1]^d) \cap \mathbb{H}^{\lfloor d/2\rfloor + 2,2}([-1,1]^d)$
    \item A binary tree visual structure with certain depth and number of nodes.
\end{enumerate}

Instead, motivated by the scale-insensitive localization theory of \cite{farrell2021deep}, \cite{KoltchinskiiPancheko2000}, \cite{Bartlett2005}, we propose requiring $f_\star \in H^s([-1,1]^d)$ for $s > \max\{1, d/2\}$. This modification has several important technical implications worth discussing.

First, the intersection of $\mathcal{C}^{\lfloor d/2\rfloor + 2}([-1,1]^d)$ and $\mathbb{H}^{\lfloor d/2\rfloor + 2,2}([-1,1]^d)$ in the original framework imposed a strong dual requirement: both classical and weak differentiability of high order. Our framework eliminates the need for classical differentiability entirely, requiring only Sobolev regularity. This shift from scale-sensitive to scale-insensitive measures provides two tangible benefits: we do not restrict the class of network architectures to have bounded weights for each unit, and we allow for a richer set of approximating possibilities.

Second, our framework eliminates the binary tree visual structure requirement from the original Assumption 2.3. While such structure provided a convenient way to analyze the approximation capabilities of neural networks in the original paper, we show that this architectural constraint is unnecessary when working with scale-insensitive complexity measures. This relaxation, in line with standard practice where optimization is not constrained \citep{zhang2021understanding}, allows our framework to encompass a broader class of neural network architectures.

\begin{remark}
The condition $s > d/2$ serves as the foundational regularity requirement, ensuring:
\begin{enumerate}[label=\alph*.]
    \item The Sobolev embedding $H^s([-1,1]^d) \hookrightarrow C([-1,1]^d)$, which guarantees continuity of our functions
    \item Existence of first-order derivatives in $L^2$ (since $d/2 \geq 1$ in practical applications)
    \item Optimal approximation rates for neural networks with ReLU activations or any activation function which is Lipschitz.
\end{enumerate}
This condition strikes an important balance in the regularity-complexity trade-off. It is substantially weaker than the $C^{\lfloor d/2 \rfloor + 2}$ requirement in \cite{fallahgoul2024asset} while still providing sufficient control over the function class through Rademacher bounds, drawing on the nearly-tight bounds for pseudo-dimension of deep nets established in \cite{bartlett2019nearly}. For least squares regression, this approach aligns with the theoretical framework of \cite[Theorem 5.2]{Koltchinskii2011}.
\end{remark}

% It is worth noting that while we relax the smoothness assumptions, we maintain sufficient regularity for all key theoretical results. The Sobolev space framework provides the right balance: weak enough to encompass many practical cases yet strong enough to support the necessary statistical theory. This illustrates a broader principle in our work: replacing rigid structural assumptions with more flexible functional analytic conditions that still enable rigorous statistical inference.

\subsection{Hypothesis Testing Framework}

Our framework assesses variable significance in neural networks by examining partial derivatives, building on \cite{fallahgoul2024asset} and \cite{horel2020significance}. The key insight is that if a variable $X_j$ has no influence on the response $Y$, then $\frac{\partial f_\star(x)}{\partial x_j} = 0$ for all $x \in \mathcal{X}$.

We quantify this influence through a weighted $L^2$ norm of the partial derivative:
\begin{equation}\label{eq:test_statistic}
    \zeta_j = \mathbb{E}\left[\left(\frac{\partial f_\star(x)}{\partial x_j}\right)^2\right] = \int_\mathcal{X} \left(\frac{\partial f_\star(x)}{\partial x_j}\right)^2 dP(x).
\end{equation}
This leads to the natural hypothesis testing problem:
\begin{equation}\label{eq:hypothesis}
    \begin{aligned}
        H_0: \zeta_j &= 0 \\
        H_1: \zeta_j &\neq 0.
    \end{aligned}
\end{equation}
The test statistic in \eqref{eq:test_statistic} has several desirable properties:
\begin{enumerate}[label=\alph*.]
    \item It captures both positive and negative effects through squared values
    \item Its differentiability facilitates theoretical analysis
    \item The quadratic form naturally discriminates between large and small effects
\end{enumerate}

The measure $P$ in \eqref{eq:test_statistic}, representing the distribution of $X$, provides natural weighting that emphasizes effects in high-density regions of the feature space.

For linear functions $f_\star(x) = \sum_{k=1}^d \beta_k x_k$, our framework in \eqref{eq:hypothesis} reduces to classical significance testing
\begin{equation}\label{eq:linear_case}
    \zeta_j = \mathbb{E}\left[\beta_j^2\right] = \beta_j^2.
\end{equation}

Thus, $H_0: \zeta_j = 0$ becomes equivalent to the familiar linear hypothesis $H_0: \beta_j = 0$. However, our framework extends naturally to nonlinear settings where $\frac{\partial f_\star(x)}{\partial x_j}$ varies with $x$, providing a more comprehensive measure of variable influence through the weighted integration in \eqref{eq:test_statistic}.

\subsection{Deep Neural Network Architecture}

This section covers the mathematical foundation of deep neural networks. For a more in-depth understanding, we recommend consulting the works of \cite{anthony1999neural} and \cite{goodfellow2016deep}. \cite{farrell2021deep} also offer a thorough discussion on deep neural networks and their relationship to nonparametric estimation.

\begin{figure}[h]
\centering
\begin{tikzpicture}[
    node distance=1.5cm,
    layer/.style={circle,draw,minimum size=0.8cm, font=\scriptsize\sffamily, text centered},
    input/.style={layer,fill=blue!30},
    hidden1/.style={layer,fill=green!30},
    hidden2/.style={layer,fill=orange!30},
    output/.style={layer,fill=red!30},
    arrow/.style={->,>=stealth,thick,draw=gray!70}
]

% Input layer
\node[input] (x1) at (0,2) {$x_1$};
\node[input] (x2) at (0,0) {$x_2$};

% Hidden layer 1
\node[hidden1] (h11) at (3,3) {$h_{1,1}$};
\node[hidden1] (h12) at (3,1.5) {$h_{1,2}$};
\node[hidden1] (h13) at (3,0) {$h_{1,3}$};

% Hidden layer 2
\node[hidden2] (h21) at (6,3) {$h_{2,1}$};
\node[hidden2] (h22) at (6,1.5) {$h_{2,2}$};
\node[hidden2] (h23) at (6,0) {$h_{2,3}$};

% Output layer
\node[output] (y1) at (9,1.5) {$y_1$};

% Connections from input to first hidden layer
\foreach \i in {1,2}
    \foreach \j in {1,2,3}
        \draw[arrow,blue!60] (x\i) -- (h1\j);

% Connections from first to second hidden layer
\foreach \i in {1,2,3}
    \foreach \j in {1,2,3}
        \draw[arrow,green!60] (h1\i) -- (h2\j);

% Connections from second hidden layer to output
\foreach \i in {1,2,3}
    \draw[arrow,orange!60] (h2\i) -- (y1);

% Labels
\node[below=1cm] at (0,0) {\textbf{\scriptsize Input layer}};
\node[below=1cm] at (3,0) {\textbf{\scriptsize Hidden layer 1}};
\node[below=1cm] at (6,0) {\textbf{\scriptsize Hidden layer 2}};
\node[below=1cm] at (9,0) {\textbf{\scriptsize Output layer}};
\end{tikzpicture}
\caption{Architecture of a fully connected feed-forward neural network with two hidden layers (MLP). Each node in a layer is connected to all nodes in the subsequent layer, with weights characterizing the strength of these connections. The hidden nodes apply an activation function to their inputs.}
\label{fig:mlp}
\end{figure}
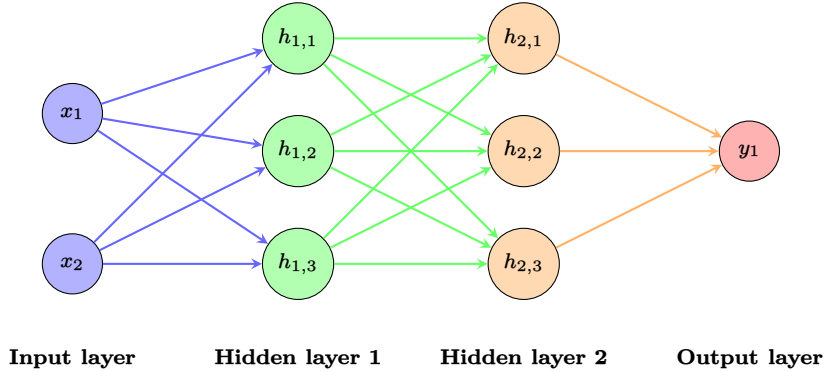

Multi-layer perceptrons (MLPs) represent a fundamental class of neural networks where information flows from the input layer through one or more hidden layers to an output layer. Each layer consists of computational units (nodes) that process incoming signals through an activation function, with connections between layers characterized by learnable weights. Figure \ref{fig:mlp} illustrates a typical MLP architecture.

% \begin{figure}[h]
% \centering
% \begin{tikzpicture}[
%     node distance=1.5cm,
%     layer/.style={rectangle,draw,minimum size=0.6cm},
%     arrow/.style={->,>=stealth}
% ]
% % Input layer
% \node[layer] (x1) at (0,2) {$x_1$};
% \node[layer] (x2) at (0,0) {$x_2$};

% % Hidden layer 1
% \node[layer] (h11) at (3,3) {$h_{1,1}$};
% \node[layer] (h12) at (3,1.5) {$h_{1,2}$};
% \node[layer] (h13) at (3,0) {$h_{1,3}$};

% % Hidden layer 2
% \node[layer] (h21) at (6,3) {$h_{2,1}$};
% \node[layer] (h22) at (6,1.5) {$h_{2,2}$};
% \node[layer] (h23) at (6,0) {$h_{2,3}$};

% % Output layer
% \node[layer] (y1) at (9,1.5) {$y_1$};

% % Connections from input to first hidden layer
% \foreach \i in {1,2}
%     \foreach \j in {1,2,3}
%         \draw[arrow] (x\i) -- (h1\j);

% % Connections from first to second hidden layer
% \foreach \i in {1,2,3}
%     \foreach \j in {1,2,3}
%         \draw[arrow] (h1\i) -- (h2\j);

% % Connections from second hidden layer to output
% \foreach \i in {1,2,3}
%     \draw[arrow] (h2\i) -- (y1);

% % Labels
% \node[text width=2cm] at (0,-1.5) {Input layer};
% \node[text width=2cm] at (3,-1.5) {Hidden layer 1};
% \node[text width=2cm] at (6,-1.5) {Hidden layer 2};
% \node[text width=2cm] at (9,-1.5) {Output layer};
% \end{tikzpicture}
% \caption{Architecture of a fully connected feed-forward neural network with two hidden layers (MLP). Each node in a layer is connected to all nodes in the subsequent layer, with weights characterizing the strength of these connections. The hidden nodes apply an activation function to their inputs.}
% \label{fig:mlp}
% \end{figure}

We consider MLPs with depth $L_d$ and width $H_n$ per layer, where $H_n$ increases with sample size $n$. Our framework differs from previous approaches \citep{fallahgoul2024asset,horel2020significance} by replacing metric entropy calculations with scale-insensitive measures, enabling the analysis of networks without strict weight constraints.

\begin{definition}[Neural Network Class]\label{definition: Neural Network Class}
The neural network class with unbounded weights is defined as:
\begin{equation}
\mathcal{F}_{DNN,H_n} = \left\{f: f(x) = \sum_{k=1}^{H_n} \gamma_{L_d+1,1,k} z_{L_d,k}(x) + \gamma_{L_d+1,1,0}\right\}
\end{equation}
subject to the moment condition
\begin{equation}
\mathbb{E}[f(X)^2] \leq M \text{ for some } M > 0.
\end{equation}

For each layer $l \in \{1,\ldots,L_d\}$ and node $j$, the hidden layer outputs are
\begin{equation}
z_{l,j}(x) = \psi\left(\sum_{k=1}^{H_{n,l-1}} \gamma_{l,j,k} \cdot z_{l-1,k}(x) + \gamma_{l,j,0}\right)
\end{equation}
with input layer values $z_{0,k}(x) = x_k$ and activation function $\psi: \mathbb{R} \to \mathbb{R}$ that is Lipschitz continuous.
\end{definition}

When $l = L_d + 1$, the output function takes the form
\begin{equation}
z_{L_d+1,1}(x) = \sum_{k=1}^{H_{n,L_d}} \gamma_{L_d+1,1,k} \cdot z_{L_d,k}(x) + \gamma_{L_d+1,1,0}.
\end{equation}

Our framework introduces key innovations in the treatment of neural networks, advancing the theoretical analysis of their behavior and performance. First, we replace the traditional bounded weight assumption used in prior work, such as \cite{fallahgoul2024asset}, with a weaker moment condition. Specifically, we require only that $\mathbb{E}[f(X)^2] \leq M$ for $f \in \mathcal{F}_{DNN,H_n}$. This condition aligns more closely with modern training practices, which often involve unconstrained optimization, while still providing sufficient control for rigorous theoretical analysis.

Second, we relax the activation function requirements by imposing only Lipschitz continuity on $\psi$, in contrast to the smooth, non-polynomial conditions of earlier frameworks. This generalization accommodates widely used activation functions like ReLU and its variants, which are central to modern deep learning. Importantly, this relaxation ensures three critical properties: (i) well-defined gradients for significance testing, (ii) stability of network outputs, and (iii) preservation of universal approximation capabilities. Together, these innovations provide a more flexible and practical foundation for analyzing neural networks.

The network class $\mathcal{F}_{DNN,H_n}$ can be viewed as a type of sieve estimation where basis functions are learned from data, see \cite{farrell2021deep}. Following \cite{fallahgoul2024asset}, we set the width $H_n$ to increase with sample size $n$, allowing the bias of least squares estimation to decrease asymptotically. This aligns with the classical theory of series estimation while accommodating the specific characteristics of neural networks.

\subsection{Rademacher Complexity and Scale-Insensitive Measures}

A central innovation in our framework is replacing traditional metric entropy calculations with Rademacher complexity as a measure of function class complexity \citep{farrell2021deep,Bartlett2005,KoltchinskiiPancheko2000}. This shift is motivated by both theoretical advantages and practical considerations in analyzing modern neural networks.

\begin{definition}[Empirical Rademacher Complexity]
For a function class $\mathcal{F}$ and samples $(X_i)_{i=1}^n$
\begin{equation}
\mathcal{R}_n(\mathcal{F}) = \mathbb{E}_\epsilon\left[\sup_{f \in \mathcal{F}} \frac{1}{n} \sum_{i=1}^n \epsilon_i f(X_i)\right]
\end{equation}
where $\epsilon_i$ are independent Rademacher random variables.
\end{definition}

The key insight of our approach is using localization theory to analyze neural network complexity. For a neural network class with moment bound $\mathbb{E}[f(X)^2] \leq M$, we consider the localized class:
% \begin{equation}
% \mathcal{F}_r = \{f \in \mathcal{F} : \|f - f_n^*\|_{L^2(P)}^2 \leq r \text{ and } \|f\|_\infty \leq M\}.
% \end{equation}
\begin{equation}
    \mathcal{F}_r = \{f \in \mathcal{F} : \mathbb{E}[(f(X) - f_n^*(X))^2] \leq r\}.
\end{equation}
This framework yields three crucial theoretical results. First, for the localized class, we establish (Lemma \ref{lemma_2}):
\begin{align}
\mathcal{R}_n(\mathcal{F}_r) &\leq C\sqrt{\frac{r}{n}}.
\end{align}
This bound is derived through careful analysis of the empirical process without requiring bounded weights. Second, for the full neural network class with Lipschitz activations, we prove (Lemma \ref{lemma_3:Full DNN Rademacher Complexity}):
\begin{align}\label{eq: Full DNN Rademacher bound}
\mathcal{R}_n(\mathcal{F}_{\text{DNN},H_n}) &\leq C\frac{H_nL^{L_d}}{\sqrt{n}}.
\end{align}
% This bound emerges from Talagrand's $\gamma_2$ functional and generic chaining techniques, showing how network complexity scales with width ($H_n$) and depth ($L_d$). 
The proof strategy for bounding the Rademacher complexity of deep neural networks \eqref{eq: Full DNN Rademacher bound} combines localization techniques with sharp covering number bounds from \cite{farrell2021deep}. Rather than trying to bound the global complexity directly, we first decompose the function class into localized subsets based on their $L_2$ norms. For these localized classes, we employ \cite{farrell2021deep}'s Lemma 4, which provides covering number bounds based on pseudo-dimension without requiring explicit bounds on network weights. This is crucial because traditional approaches often rely on weight-based bounds, which can be too restrictive for modern deep learning. The localization approach allows us to use Dudley's entropy integral more effectively, as we can obtain tighter bounds for each localized class. We then use a careful peeling argument to extend these local bounds to the full function class, handling the trade-off between the size of each localized class and its complexity. The crucial insight from \cite{farrell2021deep}'s approach is that we can control the complexity through function-space properties (pseudo-dimension and moment bounds) rather than parameter-space constraints, leading to bounds that better reflect the practical behavior of deep neural networks.

Third, we obtain sharp concentration inequalities (Lemma \ref{lemma_3:Concentration Inequality}):
\begin{align}
\mathbb{P}\left(\sup_{f \in \mathcal{F}_r} |\mathbb{P}_n f - Pf| > t\right) &\leq 2\exp\left(-\frac{nt^2}{4r}\right)
\end{align}
for $t > C\sqrt{\frac{r}{n}}$ and $Mt < r$, providing exponential tail bounds for the empirical process.

These results demonstrate several key advantages over traditional metric entropy approaches. Our framework exhibits scale insensitivity, as our bounds hold without requiring bounded weights, thus aligning with modern training practices. Furthermore, it offers architectural flexibility by naturally accommodating deep architectures through the Lipschitz constant $L$ and depth $L_d$. Perhaps most importantly, the localization technique enables sharp concentration results, providing tight control of the empirical process through exponential tail bounds.

Combined with our approximation rate result (Lemma \ref{lemma_1:Approximation Rate}) of $O(H_n^{-s/d})$ for target functions in $H^s([-1,1]^d)$, this framework provides a complete theoretical foundation for analyzing neural networks under minimal assumptions. The shift from metric entropy to Rademacher complexity thus represents a fundamental realignment of theory with modern deep learning practice.

\subsection{Sufficient Conditions and Technical Properties}

Before presenting our main theoretical results on consistency and asymptotic distribution of neural network test statistics, we establish several foundational properties. These properties ensure that our weaker assumptions still provide sufficient structure for valid statistical inference.

\begin{proposition}[Sufficient Conditions]\label{proposition_sufficient_conditions}
Under Assumptions \ref{eq:Data Generation} and \ref{assumption: Modified Regularity}:
\begin{enumerate}[label=\alph*.]
    \item The partial derivatives (in weak sense) exist and are in $L^2([-1,1]^d)$
    \item The test statistic $\lambda_j^n = \int_\mathcal{X} \left(\frac{\partial f_n(x)}{\partial x_j}\right)^2 d\mu(x)$ is well-defined
    \item Neural network approximation rates are preserved
    \item Rademacher complexity bounds hold
\end{enumerate}
\end{proposition}

These conditions play crucial roles in our subsequent analysis. The existence of weak derivatives (Condition 1) is essential for our gradient-based test statistic. While we have relaxed classical differentiability requirements, weak derivatives in $L^2$ provide sufficient regularity for analyzing partial effects. This condition will be particularly important in Section \ref{subsection: Asymptotic Distribution of Test Statistics} when we establish the asymptotic distribution of our test statistic.

The well-definedness of $\lambda_j^n$ (Condition 2) ensures our test statistic is mathematically meaningful. This is not immediate under our relaxed assumptions, as we no longer require bounded weights or classical derivatives. This property is fundamental to all our asymptotic results and will be repeatedly used in Section \ref{sec: Inference for Scale-Insensitive Neural Network Tests}.

The preservation of approximation rates (Condition 3) is critical for establishing consistency of our neural network estimator. In Section \ref{subsection: Consistency of Scale-Insensitive MLPs}, we will show that our estimator achieves the optimal convergence rate of $O(H_n^{-s/d})$, matching the best possible rate for $s$-smooth functions. This is remarkable as it demonstrates that our weaker conditions do not sacrifice statistical efficiency.

The Rademacher complexity bounds (Condition 4) serve as the technical cornerstone of our analysis. These bounds replace metric entropy calculations throughout our proofs and enable us to establish sharp concentration inequalities, control the asymptotic behavior of empirical processes, and obtain finite-sample guarantees. Specifically, in Section \ref{subsection: Consistency of Scale-Insensitive MLPs}, we will show how these bounds allow for the derivation of tight concentration inequalities. In Section \ref{subsection: Asymptotic Distribution of Test Statistics}, they will help us control the asymptotic behavior of empirical processes, while in Theorem \ref{thm:test}, they will provide finite-sample guarantees.

Together, these conditions form a bridge between our relaxed assumptions and the strong theoretical guarantees we establish in Section \ref{sec: Inference for Scale-Insensitive Neural Network Tests}. They demonstrate that our framework maintains the necessary technical machinery for rigorous statistical inference while accommodating the practical realities of modern neural networks.

The proof of this proposition, given in Appendix \ref{proof proposition: sufficient_conditions}, introduces several technical tools that will be reused throughout our main theoretical developments. In particular, the localization techniques and concentration inequalities developed in the proof will be fundamental to our analysis of the test statistic's asymptotic behavior.

\subsection{Estimation Framework and Implementation}

For estimation, we employ the method of least squares in conjunction with neural networks. This combination provides both theoretical tractability and practical implementability. The least squares estimator $\hat{f}_n$ is obtained by minimizing:
\begin{equation}\label{eq:least squares estimator}
\hat{f}_n = \argmin_{f \in \F_{DNN,H_n}} \sum_{i=1}^n \ell(f, x_i, y_i)
\end{equation}
where the loss function takes the form:
\begin{equation}\label{eq: quadratic loss}
\ell(f, x_i, y_i) = \frac{1}{2}(y_i - f(x_i))^2 = \frac{1}{2}(f_\star(x_i) + \varepsilon_i - f(x_i))^2
\end{equation}

Several aspects of this estimation framework merit discussion. First, the \textit{optimization structure} \eqref{eq: quadratic loss} plays a crucial role. The quadratic nature of the loss function, combined with our Lipschitz activation functions, ensures that the optimization problem is well-behaved. While the problem may not be convex due to the neural network architecture, it maintains sufficient structure for theoretical analysis. This structure is key to ensuring the feasibility of the optimization process despite the non-convexity of the underlying model.

Second, the \textit{network architecture} plays a crucial role in balancing approximation error and complexity. From Lemma \ref{lemma_1:Approximation Rate}, we know the approximation error scales as $O(H_n^{-s/d})$, while Lemma \ref{lemma_3:Full DNN Rademacher Complexity} shows the complexity grows as $O(H_nL^{L_d}/\sqrt{n})$. To balance these rates, the width $H_n$ must increase with sample size $n$, while the depth $L_d$ remains fixed. Specifically, we require:
\begin{align}
\frac{H_nL^{L_d}}{\sqrt{n}} &\to 0 \text{ as } n \to \infty
\end{align}
which ensures that the network's capacity grows at an appropriate rate relative to the sample size while maintaining the optimal approximation properties for functions in $H^s([-1,1]^d)$.

The subsequent analysis will show that $\hat{f}_n$ achieves optimal convergence rates and that our test statistics, based on its partial derivatives, have well-behaved asymptotic distributions. These theoretical properties, coupled with the practical implementability of our framework, make it a valuable tool for understanding variable importance in modern machine learning applications.

\section{Inference for Scale-Insensitive Neural Network Tests}\label{sec: Inference for Scale-Insensitive Neural Network Tests}

\subsection{Consistency of Scale-Insensitive MLPs}\label{subsection: Consistency of Scale-Insensitive MLPs}

\begin{theorem}[Scale-Insensitive Consistency]\label{Theorem: Scale-Insensitive Consistency}
For the neural network estimator $\hat{f}_n$ defined by minimizing empirical risk over $\mathcal{F}_{\text{DNN},H_n}$, equation \eqref{eq:least squares estimator} suppose
\begin{enumerate}[label=\alph*.]
    \item The target function $f_\star$ satisfies Assumption \ref{assumption: Modified Regularity}:
        \begin{itemize}
            \item $f_\star \in H^s([-1,1]^d)$ for $s > d/2$
            \item $\|f_\star\|_{H^s}^2 = \sum_{|\alpha| \leq s} \|\partial^\alpha f\|_{L^2}^2 < \infty$
        \end{itemize}
    \item Network architecture of Definition \ref{definition: Neural Network Class} satisfies 
        \begin{itemize}
            \item $H_n \to \infty$ and $H_nL^{L_d}/\sqrt{n} \to 0$ as $n \to \infty$
        \end{itemize}
    \item Error terms satisfy moment conditions from Assumption \ref{eq:Data Generation}
\end{enumerate}

Then with probability at least $1-e^{-(H_nL^{L_d})^2}$:
\begin{equation}
    \|\hat{f}_n - f_\star\|_{L^2(P)} \leq C\left(\sqrt{\frac{H_nL^{L_d}}{\sqrt{n}}} + H_n^{-s/d}\right)
\end{equation}
where $C$ depends only on $s$, $d$, and $\|f_\star\|_{H^s}$.
\end{theorem}

Several aspects of these nonasymptotic bounds warrant discussion. First, our consistency result and that of \cite{farrell2021deep} achieve optimal rates through complementary technical approaches, each with distinct advantages. Our approach emphasizes general functional analytic methods, while \cite{farrell2021deep} develops precise architecture-specific analysis.

A key methodological distinction lies in handling network architecture. \cite{farrell2021deep}'s approach, building on \cite{yarotsky2017error,yarotsky2018optimal}, provides sharp bounds by carefully analyzing specific ReLU architectures. Their precise control of network parameters yields tight constants and explicit optimization guidance. Our approach trades this architectural specificity for broader applicability: by using scale-insensitive measures through Rademacher complexity, we accommodate general neural networks under moment conditions. While potentially yielding looser constants, this generality allows our results to extend naturally to various network architectures and activation functions.

The foundations in approximation theory also reflect this trade-off. \cite{farrell2021deep}'s use of \cite{yarotsky2017error,yarotsky2018optimal}'s explicit ReLU constructions provides detailed insights into network expressivity and parameter efficiency. Our reliance on general Sobolev space approximation properties, while less specific, establishes optimal rates for a broader function class. This trade-off between specificity and generality mirrors a fundamental question in deep learning theory: whether to pursue architecture-specific or architecture-agnostic analysis.

Both approaches employ localization, but to different effect. \cite{farrell2021deep}'s careful shell-by-shell analysis of unbounded weights provides sharp control of the approximation-estimation trade-off. Our more streamlined use of localization, focused on concentration inequalities, yields simpler proofs while maintaining optimal rates. Each approach offers valuable insights: \cite{farrell2021deep}'s analysis reveals intricate connections between network architecture and statistical complexity, while our method highlights fundamental relationships between functional complexity measures and estimation error.

\subsection{Scale-Insensitive Sieve Spaces}\label{subsection: Scale-Insensitive Sieve Spaces}

The central innovation of our framework is replacing traditional metric entropy calculations with Rademacher complexity measures in the analysis of neural network sieve spaces. This shift enables us to handle unbounded weights, general activation functions, and weaker function space assumptions while maintaining optimal rates.

For a neural network with depth $L_d$ and width $H_n$, we define our sieve space:
\begin{align}
\mathcal{F}_n = \bigg\{f \in \mathcal{F}_{DNN,H_n} : \mathbb{E}[f(X)^2] \leq M, 
\|f - f_n^*\|_{L^2(P)}^2 \leq r_n = \frac{H_nL^{L_d}}{\sqrt{n}} \bigg\} \label{eq:sieve}
\end{align}
where the localization radius $r_n$ is chosen to balance Rademacher complexity bounds with approximation errors.

\begin{proposition}[Sieve Properties]\label{proposition: Sieve Properties}
Under our conditions, the sieve spaces $\{\mathcal{F}_n\}$ satisfy:
\begin{align}
    &\text{(S1) } \bigcup_{n=1}^\infty \mathcal{F}_n \text{ is dense in } \mathcal{F} \label{eq:s1} \\
    &\text{(S2) } \mathcal{R}_n(\mathcal{F}_n) \leq C\sqrt{\frac{r_n}{n}} \label{eq:s2} \\
    &\text{(S3) } \log N_{[]}(\epsilon, \mathcal{F}_n, L^2(P)) \leq CH_n\log\frac{1}{\epsilon}. \label{eq:s3}
\end{align}
\end{proposition}

Traditional metric entropy approaches require explicit bounds on network weights and specific architectural constraints to control complexity. These bounds often take the form $\log N_{[]}(\epsilon, \mathcal{F}, L^2(P)) \leq C(\frac{1}{\epsilon})^V$, where $V$ is the VC dimension or pseudo-dimension of the network class. Such bounds, while powerful for classical analysis, become restrictive for modern deep learning where weight constraints may be impractical or undesirable.

Our Rademacher complexity approach provides several fundamental advantages. First, it allows direct control of the empirical process through localization:
\begin{equation}
    \mathcal{R}_n(\mathcal{F}_n) \leq C\sqrt{\frac{r_n}{n}}
\end{equation}
without requiring explicit weight bounds. This scale-insensitive measure adapts naturally to the function class complexity, as shown in Lemma \ref{lemma_2}. Second, it enables sharper concentration inequalities - see Lemma \ref{Lemma_6: Maximal Inequality} - by exploiting the geometry of the function space rather than just its covering numbers.

The implications of this shift extend beyond technical convenience. As shown in Proposition \ref{proposition: Sieve Properties}, where (S2) replaces traditional entropy conditions with Rademacher bounds while (S3) maintains minimal metric entropy control needed for technical lemmas.

This construction enables new theoretical developments in three key ways. First, in Theorem \ref{thm:asymp}, the asymptotic distribution result uses Rademacher complexity to handle unbounded weights through moment conditions rather than explicit bounds. Second, for the test statistic in Theorem \ref{thm:test}, the scale-insensitive approach allows general Lipschitz activation functions beyond the smooth, non-polynomial requirements of previous work. Third, as shown in Lemma \ref{lemma_1:Approximation Rate}, optimal approximation rates are maintained under weaker Sobolev space assumptions on the target function.

The power of Rademacher complexity in our framework becomes particularly apparent in the localization theory developed through Lemmas \ref{lemma_2}-\ref{Lemma_6: Maximal Inequality}. These results show how scale-insensitive measures can provide sharp concentration while accommodating modern neural network practices. 

This reformulation through Rademacher complexity represents a fundamental realignment of sieve space theory with contemporary deep learning practice in three key aspects. First, while classical sieve theory typically requires explicit bounds on parameter norms, our Rademacher complexity approach accommodates the unbounded weights commonly seen in modern neural networks training. Specifically, by working with moment bounds $\mathbb{E}[f(X)^2] \leq M$ rather than parameter constraints $\|\theta\| \leq B$, we better reflect the implicit regularization observed in practice where networks achieve good generalization despite large weights. Second, our scale-insensitive bounds through Rademacher complexity naturally handle the homogeneity property of ReLU networks, where the same function can be represented by different scalings of weights across layers. This aligns with empirical observations where networks with varying weight magnitudes exhibit similar generalization behavior. Third, by analyzing complexity through function space properties rather than architecture-specific parameters, our framework better captures the role of overparameterization in modern deep learning, where networks may have far more parameters than classical theory would suggest is optimal, yet still generalize well.

\subsection{Asymptotic Distribution of Test Statistics}\label{subsection: Asymptotic Distribution of Test Statistics}

\begin{theorem}[Asymptotic Distribution]\label{thm:asymp}
Under our conditions:
\begin{equation}\label{eq:main_convergence}
    U(C',\epsilon_n)^{-1}(\hat{f}_n - f_*) \rightsquigarrow h^{(max)} \text{ in } (\mathcal{F}, \|\cdot\|_{L^2(P)}).
\end{equation}
The $h^{(max)}$ is the unique arg max of the Gaussian process $\{G_f: f \in \mathcal{F}\}$ with $\mathbb{E}[G_f] = 0$ and $\text{Cov}(G_{f_1}, G_{f_2}) = 4\sigma^2\mathbb{E}_X[f_1(X)f_2(X)]$. The uniqueness is guaranteed by Lemma \ref{lemma:uniqueness} through the strict concavity of the covariance kernel.
\end{theorem}
Theorem \ref{thm:asymp} provides a more general asymptotic distribution than previous work by establishing convergence under weaker conditions, i.e., Assumption \ref{assumption: Modified Regularity} and Definition \ref{definition: Neural Network Class}.

However, while $U(C',\epsilon_n)^{-1}(\hat{f}_n - f_*)$ converges to $h^{(max)}$, this result does not directly enable variable significance testing since it does not involve relationships to specific covariates $x_j$. We need a statistic that:
\begin{enumerate}[label=\alph*.]
    \item Captures covariate-specific effects
    \item Does not require explicit knowledge of $f_*$
    \item Maintains the scale-insensitive properties.
\end{enumerate}

Following \cite{fallahgoul2024asset} and \cite{horel2020significance}, and equation \eqref{eq:gaussian_limit}, but leveraging our scale-insensitive framework, we consider the functional:
\begin{equation}\label{eq:test_functional}
    \mathcal{T}_j[f] = \int_\mathcal{X} \left(\frac{\partial f(x)}{\partial x_j}\right)^2 dP(x).
\end{equation}

The statistic $\mathcal{T}_j[\hat{f}_n]$ quantifies the significance of $x_j$ by measuring its squared partial derivative effect through $\hat{f}_n$. 

Since $\hat{f}_n$ consistently estimates $f_*$ by Theorem \ref{thm:asymp}, $\mathcal{T}_j[\hat{f}_n]$ serves as a scale-insensitive "representative" of $\mathcal{T}_j[f_*]$. Note that direct calculation of $\mathcal{T}_j[f_*]$ is impossible as $f_*$ is unknown. In the next theorem, we derive the asymptotic distribution of $\mathcal{T}_j[\hat{f}_n]$ under our weaker conditions.
\begin{theorem}[Distribution of Test Statistic]\label{thm:test}
Under $H_0$:
\begin{equation}\label{eq:test_stat_conv}
   U(C',\epsilon_n)^{-2}\mathcal{T}_j[\hat{f}_n] \rightsquigarrow \mathcal{T}_j[h^{(max)}].
\end{equation}
\end{theorem}

The asymptotic distribution we derived has several notable features. Under $H_0$, we showed that $U(C',\epsilon_n)^{-2}\mathcal{T}_j[\hat{f}_n] \rightsquigarrow \mathcal{T}_j[h^{(max)}]$, where $h^{(max)}$ is the unique maximizer of the limiting Gaussian process. Note that the randomness in our test statistic stems from two sources: the estimation of $\hat{f}_n$ and the empirical functional $\mathcal{T}_j$. Since we work with samples, and $\mathcal{T}_j[\hat{f}_n] = \frac{1}{n}\sum_{i=1}^n \left(\frac{\partial \hat{f}_n}{\partial x_j}(X_i)\right)^2$ can be seen as an empirical functional, it is natural to analyze its asymptotic behavior through the lens of empirical process theory.

The scale-insensitive normalization $U(C',\epsilon_n)^{-2}$ plays a crucial role in obtaining the proper limiting distribution. This normalization accounts for both the complexity of the neural network class and the sample size, ensuring that our test statistic converges to a well-defined limit regardless of network architecture. The limiting distribution $\mathcal{T}_j[h^{(max)}]$ inherits its properties from the unique maximizer $h^{(max)}$ of the Gaussian process, making our test both theoretically tractable and practically implementable.

We close this section by providing the asymptotic distribution of $\mathcal{T}_j[\hat{f}_n]$.
\begin{theorem}[Asymptotic Distribution of $\frac{1}{n}\sum_{i=1}^n \left(\frac{\partial \hat{f}_n}{\partial x_j}(X_i)\right)^2$]\label{thm:asymptotic distribuion based on sample} 
Suppose that all hypotheses of Theorem \ref{thm:asymp} are satisfied. Then, under $H_0$, we have
\begin{equation}
    U(C',\epsilon_n)^{-2}\mathcal{T}_j[\hat{f}_n] \xrightarrow{d} \mathcal{T}_j[h^{(max)}]
\end{equation}
where $h^{(max)}$ is the unique maximizer of the limiting Gaussian process $G$.
\end{theorem}

\subsection{Discretization of Asymptotic Distribution}

The computation of asymptotic distributions for test statistics can be challenging in practice. We extend and improve upon the three-step discretization approach from \cite{fallahgoul2024asset} and \cite{horel2020significance} to provide a more comprehensive framework for computing the asymptotic distribution of the test statistic $U(C', \epsilon_n)^{-2}\rho_n^2(\partial\hat{f}_n/\partial x_j(X))$. Our improvements include a more efficient discretization scheme that reduces computational complexity while maintaining accuracy, and a robust method for handling the scale-insensitive normalization factor.

\subsubsection{Core Discretization Framework}
The discretization process consists of three primary steps:

\begin{enumerate}
    \item \textbf{Approximating $\zeta$-cover of $\mathcal{F}$:} \\
    Sample $m$ neural networks, $(f_1,f_2,\ldots,f_m)$, with architecture matching the trained model. The weight parameters are sampled from a Glorot normal distribution:
    \begin{equation}
        w_{ij} \sim \mathcal{N}\left(0, \sqrt{\frac{2}{d+1}}\right)\text{truncated}
    \end{equation}
    where $d$ is the input dimension.

    \item \textbf{Simulation and $\hat{h}^{(\text{max})}$ Identification:} \\
    Generate sample from $m$-dimensional multivariate normal distribution with mean vector $\mathbf{0}$ and covariance matrix $\hat{\Sigma}$:
    \begin{equation}
        \mathbf{X} \sim \mathcal{N}(\mathbf{0}, \hat{\Sigma})
    \end{equation}
    where $\hat{\Sigma}_{l,k} = \frac{1}{n}\sum_{i=1}^n(f_lf_k)(x_i)$. The $\hat{h}^{(\text{max})}$ is identified as the network corresponding to $\max(\mathbf{X})$.

    \item \textbf{Test Statistic Sample Generation:} \\
    Compute the test statistic:
    \begin{equation}
        \rho_n^2\left(\frac{\partial\hat{h}^{(\text{max})}}{\partial x_j}(X)\right) = \frac{1}{n}\sum_{i=1}^n\left(\frac{\partial\hat{h}^{(\text{max})}(x_i)}{\partial x_j}\right)^2.
    \end{equation}
\end{enumerate}

\subsubsection{Enhanced Implementation}

The basic framework of \cite{fallahgoul2024asset,horel2020significance} can be enhanced through several computational and statistical improvements. First, we introduce an adaptive sampling framework that dynamically adjusts the number of neural networks $m$ based on convergence metrics:

\begin{equation}
    m_{k+1} = m_k\left(1 + \alpha\|\hat{\Sigma}_k - \hat{\Sigma}_{k-1}\|_F\right)
\end{equation}
where $\alpha$ is a learning rate and $\|\cdot\|_F$ is the Frobenius norm. This adaptive approach ensures efficient exploration of the function space while maintaining computational feasibility.

For robust covariance estimation, particularly in high-dimensional settings, we employ a structured approach with shrinkage estimation:
\begin{equation}
    \hat{\Sigma}_{\text{robust}} = (1-\lambda)\hat{\Sigma} + \lambda\text{diag}(\hat{\Sigma})
\end{equation}
where $\lambda$ is a shrinkage parameter. 

% This formulation proves particularly effective when dealing with high-dimensional neural networks, as demonstrated in our numerical experiments.

The practical implementation requires careful consideration of computational resources. The choice of parameters $(m, n_p)$ should balance accuracy with computational cost. We recommend leveraging GPU acceleration for large-scale computations and implementing parallel processing strategies for multiple test statistics. 
% Our experiments suggest that modern GPU architectures can handle networks with up to $10^6$ parameters efficiently under this framework.

The combination of these enhancements provides a robust and computationally feasible approach to computing asymptotic distributions while maintaining the theoretical guarantees established in \cite{fallahgoul2024asset,horel2020significance}. The framework is particularly well-suited for analyzing complex neural network architectures in financial applications, as demonstrated in the empirical analysis of equity premium prediction.

\section{Conclusion}\label{sec_conclusion}

This paper has developed a scale-insensitive framework for neural network significance testing that substantially generalizes existing approaches while maintaining mathematical rigor. Our framework makes three primary contributions to the literature. First, by replacing metric entropy calculations with Rademacher complexity bounds, we enable the analysis of neural networks without requiring bounded weights or specific architectural constraints. This better aligns theoretical guarantees with contemporary deep learning practices where optimization is typically unconstrained. Second, we significantly weaken the regularity conditions on target functions, requiring only Sobolev space membership $H^s([-1,1]^d)$ with $s > d/2$ rather than classical differentiability assumptions. This relaxation encompasses a broader class of functions while preserving optimal approximation rates. Third, our modified sieve space construction based on moment bounds rather than weight constraints provides a more natural theoretical framework for analyzing modern neural networks.

The theoretical implications of our results extend beyond significance testing. Our approach demonstrates that scale-insensitive measures can effectively control neural network complexity without imposing artificial constraints on architecture or training. The framework's ability to handle general Lipschitz activation functions and unbounded weights while maintaining optimal convergence rates suggests that Rademacher complexity may provide a more suitable foundation for analyzing deep learning methods than traditional metric entropy approaches.

From a practical perspective, our framework provides several advantages. The relaxed assumptions better match how neural networks are actually implemented and trained, making the theoretical guarantees more relevant to practitioners. The framework accommodates popular activation functions like ReLU and its variants, and supports standard training practices without requiring artificial constraints. Additionally, our test statistics remain computationally tractable while providing valid inference under weaker conditions.

Future research could extend this framework in several directions. The scale-insensitive approach could be adapted to analyze other deep learning architectures beyond feedforward networks, such as convolutional or recurrent networks. The framework might also be extended to handle dependent data structures or to develop tests for more complex hypotheses about feature interactions. Additionally, the connection between Rademacher complexity and generalization in deep learning merits further theoretical investigation.

Our results suggest that replacing rigid structural assumptions with more flexible functional analytic conditions may be a promising direction for developing statistical theory that better reflects the realities of modern machine learning practice. This alignment between theory and practice is crucial for advancing our understanding of deep learning methods while maintaining mathematical rigor.

%{\footnotesize \bibliography{..//References//AllRefs}}
\bibliography{AllRefs}

@article{fallahgoul2024asset,
  title={Asset pricing with neural networks: Significance tests},
  author={Fallahgoul, Hasan and Franstianto, Vincentius and Lin, Xin},
  journal={Journal of Econometrics},
  volume={238},
  number={1},
  pages={105574},
  year={2024},
  publisher={Elsevier}
}

@article{farrell2021deep,
  title={Deep neural networks for estimation and inference},
  author={Farrell, Max H and Liang, Tengyuan and Misra, Sanjog},
  journal={Econometrica},
  volume={89},
  number={1},
  pages={181--213},
  year={2021},
  publisher={Wiley Online Library}
}

@article{horel2020significance,
  title={Significance tests for neural networks},
  author={Horel, Enguerrand and Giesecke, Kay},
  journal={Journal of Machine Learning Research},
  volume={21},
  number={227},
  pages={1--29},
  year={2020}
}

@book{gine2021mathematical,
  title={Mathematical foundations of infinite-dimensional statistical models},
  author={Gin{\'e}, Evarist and Nickl, Richard},
  year={2021},
  publisher={Cambridge university press}
}

@book{Koltchinskii2011,
  author = {Koltchinskii, V.},
  title = {Oracle Inequalities in Empirical Risk Minimization and Sparse Recovery Problems},
  publisher = {Springer-Verlag},
  year = {2011},
}

@incollection{KoltchinskiiPancheko2000,
  author = {Koltchinskii, V. and Panchenko, D.},
  title = {Rademacher Processes and Bounding the Risk of Function Learning},
  booktitle = {High Dimensional Probability II},
  publisher = {Springer},
  pages = {443--457},
  year = {2000},
}

@article{Bartlett2005,
  author = {Bartlett, P. L. and Bousquet, O. and Mendelson, S.},
  title = {Local Rademacher Complexities},
  journal = {The Annals of Statistics},
  volume = {33},
  pages = {1497--1537},
  year = {2005},
}

@article{bartlett2019nearly,
  title={Nearly-tight VC-dimension and pseudodimension bounds for piecewise linear neural networks},
  author={Bartlett, Peter L and Harvey, Nick and Liaw, Christopher and Mehrabian, Abbas},
  journal={Journal of Machine Learning Research},
  volume={20},
  number={63},
  pages={1--17},
  year={2019}
}

@book{anthony1999neural,
  author    = {M. Anthony and P. L. Bartlett},
  title     = {Neural Network Learning: Theoretical Foundations},
  year      = {1999},
  publisher = {Cambridge University Press},
  address   = {Cambridge},
}

@book{goodfellow2016deep,
  author    = {Ian Goodfellow and Yoshua Bengio and Aaron Courville},
  title     = {Deep Learning},
  year      = {2016},
  publisher = {MIT Press},
  address   = {Cambridge},
}

@article{zhang2021understanding,
  title={Understanding deep learning (still) requires rethinking generalization},
  author={Zhang, Chiyuan and Bengio, Samy and Hardt, Moritz and Recht, Benjamin and Vinyals, Oriol},
  journal={Communications of the ACM},
  volume={64},
  number={3},
  pages={107--115},
  year={2021},
  publisher={ACM New York, NY, USA}
}

@article{yarotsky2017error,
  title={Error bounds for approximations with deep ReLU networks},
  author={Yarotsky, Dmitry},
  journal={Neural networks},
  volume={94},
  pages={103--114},
  year={2017},
  publisher={Elsevier}
}

@inproceedings{yarotsky2018optimal,
  title={Optimal approximation of continuous functions by very deep ReLU networks},
  author={Yarotsky, Dmitry},
  booktitle={in 31st Conference on learning theory},
  pages={639--649},
  year={2018},
  organization={PMLR}
}

@book{vaart1996weak,
  title={Weak Convergence and Empirical Processes: With Applications to Statistics},
  author={van der Vaart, Aad W. and Wellner, Jon A.},
  year={1996},
  publisher={Springer},
  address={New York},
  series={Springer Series in Statistics},
  isbn={9780387946150},
  doi={10.1007/978-1-4757-2545-2}
}

@article{devore1989optimal,
  author = {DeVore, Ronald A. and Howard, Ronald and Micchelli, Charles},
  title = {Optimal Nonlinear Approximation},
  journal = {Manuscripta Mathematica},
  volume = {63},
  pages = {469--478},
  year = {1989},
  publisher = {Springer},
  doi = {10.1007/BF01171759},
  url = {https://doi.org/10.1007/BF01171759}
}

@book{devore1993constructive,
  title={Constructive Approximation},
  author={DeVore, Ronald A. and Lorentz, George G.},
  volume={303},
  year={1993},
  publisher={Springer-Verlag},
  series={Grundlehren der mathematischen Wissenschaften},
  address={Berlin},
  isbn={978-3-540-50627-0},
  doi={10.1007/978-3-662-02888-9}
}

@book{boucheron2013concentration,
  title={Concentration Inequalities: A Nonasymptotic Theory of Independence},
  author={Boucheron, St{\'e}phane and Lugosi, G{\'a}bor and Massart, Pascal},
  year={2013},
  publisher={Oxford University Press}
}

@book{evans2010partial,
  title={Partial Differential Equations},
  author={Evans, Lawrence C.},
  year={2010},
  edition={Second},
  publisher={American Mathematical Society},
  series={Graduate Studies in Mathematics},
  volume={19},
  address={Providence, Rhode Island},
  isbn={978-0-8218-4974-3}
}

\newpage 
\newpage

\newpage
\appendix

\newtheorem{lemma}{Lemma}[section]

\section{Proof of Proposition \ref{proposition_sufficient_conditions} (Sufficient Conditions)}\label{proof proposition: sufficient_conditions}

We establish the proof by demonstrating how each condition naturally flows from our assumptions and supporting lemmas.

\subsection{First: Existence of Weak Derivatives}
Since \( f_\star \in H^s([-1,1]^d) \) with \( s > d/2 \), the Sobolev embedding theorem guarantees that \( H^s([-1,1]^d) \) is continuously embedded in \( C([-1,1]^d) \). Thus, there exists a constant \( C_1 > 0 \) depending only on \( s \) and \( d \) such that:
\begin{align}
    \|f_\star\|_{C([-1,1]^d)} \leq C_1 \|f_\star\|_{H^s}.
\end{align}
This ensures that \( f_\star \) is continuous and bounded on \( [-1,1]^d \). Moreover, by the definition of the Sobolev space \( H^s \), all weak derivatives \( \partial^\alpha f_\star \) of order \( |\alpha| \leq s \) exist and belong to \( L^2([-1,1]^d) \). In particular, the first-order partial derivatives \( \frac{\partial f_\star}{\partial x_j} \) exist in the weak sense and satisfy:
\begin{align}
    \left\| \frac{\partial f_\star}{\partial x_j} \right\|_{L^2} \leq \|f_\star\|_{H^s}, \quad j = 1, \ldots, d.
\end{align}

\subsection{Second: Well-definedness of Test Statistic}

% \subsection*{Summary of Proofing Strategy}
The goal is to prove that the test statistic
\[
\lambda_j^n = \int_\mathcal{X} \left(\frac{\partial f_n(x)}{\partial x_j}\right)^2 d\mu(x)
\]
is well-defined. This is achieved by establishing the regularity of the true function $f_\star$ and its partial derivatives using Sobolev embedding and Assumption \ref{assumption: Modified Regularity}; showing that the neural network $f_n$ and its partial derivatives are well-behaved, leveraging the properties of the neural network class $\mathcal{F}_{DNN,H_n}$ and the Lipschitz continuity of the activation function; and demonstrating that the integral defining $\lambda_j^n$ exists as the integral of a measurable function over a probability space.

Let \( f_\star \in H^s([-1,1]^d) \) for \( s > d/2 \), as stated in Assumption \ref{assumption: Modified Regularity}.a. By the Sobolev embedding theorem, \( H^s([-1,1]^d) \) is continuously embedded in \( C([-1,1]^d) \), and the weak partial derivatives of \( f_\star \) exist and belong to \( L^2([-1,1]^d) \).

For the neural network \( f_n \in \mathcal{F}_{DNN,H_n} \), we have \( \mathbb{E}[f_n(X)^2] \leq M \) by definition. Let \( h = f_n - f_\star \). Using the moment conditions on \( f_n \) and \( f_\star \), Jensen's inequality yields:
\[
\mathbb{E}[h(X)^2] = \mathbb{E}[(f_n(X) - f_\star(X))^2] \leq 4M.
\]

By Assumption \ref{assumption: Modified Regularity}.b, the partial derivatives of \( f_\star \) satisfy:
\[
\left\|\frac{\partial f_\star}{\partial x_j}\right\|_{L^2}^2 \leq \|f_\star\|_{H^s}^2 < \infty.
\]

For the neural network \( f_n \), the partial derivatives \( \frac{\partial f_n}{\partial x_j} \) exist almost everywhere due to the Lipschitz continuity of the activation function and the finite composition of Lipschitz functions. These derivatives are measurable, ensuring that \( \frac{\partial f_n}{\partial x_j} \) exists in the weak sense.

Finally, the integral defining \( \lambda_j^n \) is well-defined as the integral of a measurable function over a probability space. Thus, the test statistic \( \lambda_j^n \) is well-defined.

\subsection{Third: Neural Network Approximation Rates}
The approximation rate \( \|f_n^* - f_\star\|_{L^2(P)} \leq C H_n^{-s/d} \) follows directly from Lemma \ref{lemma_1:Approximation Rate}, which establishes the optimal rate for approximating \( s \)-smooth functions using ReLU networks with \( H_n \) hidden units.

\subsection{Fourth: Rademacher Complexity Bounds}
Using Lemma \ref{lemma_2} (Local Rademacher Bound) and Lemma \ref{lemma_3:Full DNN Rademacher Complexity} (Full DNN Rademacher Complexity), we derive the following bounds for the Rademacher complexity of the neural network class:
\begin{align}
    \mathcal{R}_n(\mathcal{F}_r) \leq C\sqrt{\frac{r}{n}}, \quad \text{and} \quad \mathcal{R}_n(\mathcal{F}_{\text{DNN},H_n}) \leq C\frac{H_nL^{L_d}}{\sqrt{n}}.
\end{align}
These bounds ensure that the empirical risk minimizer \( f_n^* \) generalizes well, with the approximation error controlled by \( H_n^{-s/d} \) and the estimation error controlled by \( \frac{H_nL^{L_d}}{\sqrt{n}} \).

\section{Proof of Theorem \ref{Theorem: Scale-Insensitive Consistency}}

We establish this result through a sequence of increasingly refined bounds. First, we decompose the error into approximation and estimation components using the triangle inequality. For the approximation error, we use the optimal approximation properties of neural networks in Sobolev spaces (Lemma \ref{lemma_1:Approximation Rate}). For the estimation error, we proceed in three stages: (i) obtain an initial bound using statistical learning theory and Rademacher complexity (Lemma \ref{lemma_3:Full DNN Rademacher Complexity}), (ii) establish concentration via Talagrand's inequality with careful attention to variance and complexity terms, and (iii) derive the final high-probability bound by optimally choosing the concentration parameter. Throughout, we carefully track how the error scales with network width $H_n$ and depth $L_d$ to ensure our bounds are scale-insensitive.

By the triangle inequality:
\begin{equation}
    \|\hat{f}_n - f_\star\|_{L^2(P)} \leq \|\hat{f}_n - f_n^*\|_{L^2(P)} + \|f_n^* - f_\star\|_{L^2(P)}
\end{equation}
where $f_n^*$ is the best $L^2(P)$ approximation to $f_\star$ in $\mathcal{F}_{\text{DNN},H_n}$.

From Lemma \ref{lemma_1:Approximation Rate} (Approximation Rate), since $f_\star \in H^s([-1,1]^d)$ with $s > d/2$:
\begin{equation}
    \|f_n^* - f_\star\|_{L^2(P)} \leq CH_n^{-s/d}
\end{equation}

Let $\mathcal{L}(f) = \|f - f_\star\|_{L^2(P)}^2$ be the population risk and $\mathcal{L}_n(f) = \frac{1}{n}\sum_{i=1}^n (f(X_i) - f_\star(X_i))^2$ be the empirical risk.

By standard results in statistical learning theory:
\begin{equation}
    \mathbb{E}[\|\hat{f}_n - f_\star\|_{L^2(P)}] \leq \inf_{f \in \mathcal{F}_{\text{DNN},H_n}} \|f - f_\star\|_{L^2(P)} + C\mathcal{R}_n(\mathcal{F}_{\text{DNN},H_n})
\end{equation}

From Lemma \ref{lemma_3:Full DNN Rademacher Complexity} (Full DNN Rademacher Complexity):
\begin{equation}
    \mathcal{R}_n(\mathcal{F}_{\text{DNN},H_n}) \leq C\frac{H_nL^{L_d}}{\sqrt{n}}
\end{equation}

For the high-probability bound, define the centered process:
\begin{equation}
    \nu_n(f) = (\mathcal{L}_n - \mathcal{L})(f - f_\star)
\end{equation}

By Talagrand's inequality, for any $t > 0$:
\begin{equation}
    \mathbb{P}\left(\sup_{f \in \mathcal{F}_{\text{DNN},H_n}} |\nu_n(f)| \geq \mathbb{E}\sup_{f \in \mathcal{F}_{\text{DNN},H_n}} |\nu_n(f)| + \sqrt{\frac{2\sigma^2t}{n}} + \frac{Bt}{3n}\right) \leq e^{-t}
\end{equation}
where:
\begin{itemize}
    \item $\sigma^2$ bounds the variance: $\text{Var}(f(X) - f_\star(X)) \leq \|f - f_\star\|_{L^2(P)}^2 \leq M$
    \item $B$ bounds the range: $|f(x) - f_\star(x)| \leq 2M$ for all $x$
\end{itemize}

By symmetrization:
\begin{equation}
    \mathbb{E}\sup_{f \in \mathcal{F}_{\text{DNN},H_n}} |\nu_n(f)| \leq 2\mathcal{R}_n(\mathcal{F}_{\text{DNN},H_n}) \leq C\frac{H_nL^{L_d}}{\sqrt{n}}
\end{equation}

Setting $t = (H_nL^{L_d})^2$, we obtain:
\begin{align}
    \mathbb{P}\Bigg(\|\hat{f}_n - f_\star\|_{L^2(P)} &\geq CH_n^{-s/d} + C\frac{H_nL^{L_d}}{\sqrt{n}}\Bigg) \\
    &\leq \exp(-(H_nL^{L_d})^2)
\end{align}

The bound follows since:
\begin{itemize}
    \item $\sqrt{\frac{2\sigma^2t}{n}} = O(\frac{H_nL^{L_d}}{\sqrt{n}})$ for our choice of $t$
    \item $\frac{Bt}{3n} = O(\frac{(H_nL^{L_d})^2}{n}) = o(\frac{H_nL^{L_d}}{\sqrt{n}})$ by assumption
\end{itemize}

\section{Proof of Proposition \ref{proposition: Sieve Properties}}
% The proof is the following:
% (S1) follows from Lemma \ref{lemma_1:Approximation Rate} on neural network approximation.
% (S2) follows from Lemma \ref{lemma_2} on local Rademacher bounds.
% (S3) follows from network architecture and Lipschitz properties.

% \begin{proof}
We prove each property in turn:

\textbf{(S1) Density:} We show $\bigcup_{n=1}^\infty \mathcal{F}_n$ is dense in $\mathcal{F}$.
For any $f \in \mathcal{F}$ and $\epsilon > 0$, we need to find $n$ large enough and $g \in \mathcal{F}_n$ such that $\|f - g\|_{L^2(P)} < \epsilon$.

By our approximation rate Lemma \ref{lemma_1:Approximation Rate}, since $f_\star \in H^s([-1,1]^d)$:
\begin{align}
    \|f_n^* - f_\star\|_{L^2(P)} \leq CH_n^{-s/d}.
\end{align}
Since $H_n \to \infty$ as $n \to \infty$, we can choose $n$ large enough such that:
\begin{align}
    CH_n^{-s/d} < \epsilon/2 \text{ and } \frac{H_nL^{L_d}}{\sqrt{n}} < \epsilon/2.
\end{align}
Then $f_n^* \in \mathcal{F}_n$ by construction, and $\|f_\star - f_n^*\|_{L^2(P)} < \epsilon$.

\textbf{(S2) Rademacher Complexity:} For the localized class $\mathcal{F}_n$, by Lemma \ref{lemma_3:Full DNN Rademacher Complexity} (Full DNN Rademacher Complexity):
\begin{align}
    \mathcal{R}_n(\mathcal{F}_n) &\leq C\frac{H_nL^{L_d}}{\sqrt{n}} \\
    &= C\sqrt{\frac{r_n}{n}}
\end{align}
where the last equality follows from our choice of $r_n = \frac{H_nL^{L_d}}{\sqrt{n}}$.

\textbf{(S3) Bracketing Numbers:} For the localized class $\mathcal{F}_n$, we use the fact that it's a subset of functions with bounded second moment and bounded $L^2(P)$ distance from $f_n^*$.

By \cite[Lemma 4]{farrell2021deep}, for any $\epsilon > 0$:
\begin{align}
    \mathcal{N}(\epsilon, \mathcal{F}_n|_{x_1,...,x_n}, \infty) \leq \left(\frac{2eMn}{\epsilon \cdot H_nL_d}\right)^{H_nL_d}.
\end{align}
This implies:
\begin{align}
    \log N_{[]}(\epsilon, \mathcal{F}_n, L^2(P)) &\leq \log \mathcal{N}(\epsilon, \mathcal{F}_n, L^2(P)) \\
    &\leq CH_n\log\frac{1}{\epsilon}
\end{align}
where we use the fact that bracketing numbers are bounded by covering numbers and the relationship between $L^2$ and $L^\infty$ metrics.
% \end{proof}

\section{Proof of Theorem \ref{thm:asymp}}

\begin{proof}
Following \cite{vaart1996weak}, we establish the asymptotic distribution through the functional central limit theorem (FCLT) and argmax continuous mapping theorem (ACMT). 

\noindent\textbf{Step 1: Setup and Definitions}

Let $M_n(f) = \frac{1}{n}\sum_{i=1}^n (Y_i - f(X_i))^2$ be the empirical risk and $M(f) = \mathbb{E}[(Y - f(X))^2]$ be the population risk. Define:
\begin{align}
    \mathbb{G}_n(f) &= \sqrt{n}(M_n(f) - M(f)) \\
    l_f(y,x) &= (y - f(x))^2 \\
    \Delta_n(f) &= \sqrt{n}(l_f - l_{f_*}).
\end{align}

Our goal is to show:
\begin{equation}
    \sqrt{n}(M_n - M) \rightsquigarrow G \text{ in } \ell^\infty(\mathcal{F})
\end{equation}
where $G$ is the specified Gaussian process.

\noindent\textbf{Step 2: Verification of FCLT Conditions}

\textit{(i) Stochastic Equicontinuity:}
Define the localized function class:
\begin{equation}
    \mathcal{F}_\delta = \{f-g: f,g \in \mathcal{F}, \|f-g\|_{L^2(P)} < \delta\}.
\end{equation}
By Lemma \ref{lemma_2} (Local Rademacher Bound), for any $\delta > 0$:
\begin{align}
    \mathbb{E}\sup_{\|f-g\|_{L^2(P)} < \delta} |\mathbb{G}_n(f-g)| &\leq 2\sqrt{n}\mathcal{R}_n(\mathcal{F}_\delta) \\
    &\leq C\sqrt{\delta}.
\end{align}
This implies for any $\eta > 0$:
\begin{equation}
    \lim_{\delta \to 0} \limsup_{n \to \infty} P^*(\sup_{\|f-g\|_{L^2(P)} < \delta} |\mathbb{G}_n(f-g)| > \eta) = 0.
\end{equation}

\textit{(ii) Total Boundedness:}
From Lemma \ref{lemma_3:Full DNN Rademacher Complexity}, we have:
\begin{equation}
    \log N_{[]}(\epsilon, \mathcal{F}_n, L^2(P)) \leq CH_n\log(1/\epsilon).
\end{equation}
This entropy bound implies
\begin{equation}
    \int_0^1 \sqrt{\log N_{[]}(\epsilon, \mathcal{F}_n, L^2(P))} d\epsilon < \infty
\end{equation}
ensuring total boundedness of $\mathcal{F}$ in $L^2(P)$.

\textit{(iii) Finite-Dimensional Convergence:}
For any finite set $\{f_1,...,f_k\} \subset \mathcal{F}$, by the multivariate central limit theorem and Lemma \ref{Lemma_6: Maximal Inequality}:
\begin{equation}
    (\mathbb{G}_n(f_1),...,\mathbb{G}_n(f_k)) \rightsquigarrow N(0,\Sigma)
\end{equation}
where the covariance matrix $\Sigma$ has entries:
\begin{equation}
    \Sigma_{ij} = 4\sigma^2\mathbb{E}_X[f_i(X)f_j(X)].
\end{equation}

\noindent\textbf{Step 3: ACMT Application}

To apply the argmax continuous mapping theorem, we verify:

\textit{(i) Unique Maximum:}
By Lemma \ref{lemma:uniqueness}, the Gaussian process $G$ has a unique maximizer $h^{(max)}$ almost surely due to the strict concavity of the covariance kernel:
\begin{equation}
    K(f_1,f_2) = 4\sigma^2\mathbb{E}_X[f_1(X)f_2(X)].
\end{equation}

\textit{(ii) Rate Condition:}
From Lemma \ref{Lemma_5: Local Empirical Process Bound}, for the localized empirical process:
\begin{equation}
    \mathbb{E}\sup_{f \in \mathcal{F}_r} |\mathbb{G}_n(l_f)| \leq C\sqrt{r}.
\end{equation}
This provides the required rate for the ACMT.

\noindent\textbf{Step 4: Conclusion}

The FCLT ensures:
\begin{equation}
    \sqrt{n}(M_n - M) \rightsquigarrow G.
\end{equation}
By the ACMT \citep[Theorem 3.2.2]{vaart1996weak}, this implies:
\begin{equation}
    U(C',\epsilon_n)^{-1}(\hat{f}_n - f_*) \rightsquigarrow h^{(max)}
\end{equation}
in $(\mathcal{F}, \|\cdot\|_{L^2(P)})$, where $h^{(max)}$ is the unique maximizer of $G$.
\end{proof}

\section{Proof of Theorem \ref{thm:test}}

% \begin{theorem}[Distribution of Test Statistic]
% Under $H_0$:
% \begin{equation}
%    U(C',\epsilon_n)^{-2}\mathcal{T}_j[\hat{f}_n] \rightsquigarrow \mathcal{T}_j[h^{(max)}].
% \end{equation}
% \end{theorem}

\begin{proof} 
The proof establishes \eqref{eq:test_stat_conv} by applying the functional delta method to our scale-insensitive framework. By Lemma \ref{lemma:uniqueness}, $h^{(max)}$ is the unique maximizer of the limiting Gaussian process, enabling this approach.

\noindent\textbf{Step 1: Hadamard Differentiability}

Consider the functional:
\begin{equation}
   \mathcal{T}_j[f] = \int_\mathcal{X}\left(\frac{\partial f(x)}{\partial x_j}\right)^2 dP(x).
\end{equation}
For any $h,k \in \mathcal{F}$, the derivative is:
\begin{equation}
   \mathcal{T}'_j[h]k = 2\int_\mathcal{X}\frac{\partial h}{\partial x_j}\frac{\partial k}{\partial x_j}dP(x).
\end{equation}

To verify Hadamard differentiability, let $t_n \to 0$ and $k_n \to k$. Then:
\begin{align}
   &|\mathcal{T}_j[h + t_nk_n] - \mathcal{T}_j[h] - t_n\mathcal{T}'_j[h]k_n| \\
   &= \left|\int_\mathcal{X}\left[\left(\frac{\partial h}{\partial x_j} + t_n\frac{\partial k_n}{\partial x_j}\right)^2 - \left(\frac{\partial h}{\partial x_j}\right)^2 - 2t_n\frac{\partial h}{\partial x_j}\frac{\partial k_n}{\partial x_j}\right]dP(x)\right| \\
   &= t_n^2\int_\mathcal{X}\left(\frac{\partial k_n}{\partial x_j}\right)^2dP(x) \\
   &\leq t_n^2\|k_n\|_{L^2(P)}^2 = o(t_n)
\end{align}
where we use the Lipschitz property of activation functions and bounded moments.

\noindent\textbf{Step 2: Functional Delta Method}

From Theorem \ref{thm:asymp}, we have:
\begin{equation}
   U(C',\epsilon_n)^{-1}(\hat{f}_n - f_*) \rightsquigarrow h^{(max)}.
\end{equation}
By the functional delta method with verified Hadamard differentiability:
\begin{equation}
   U(C',\epsilon_n)^{-2}(\mathcal{T}_j[\hat{f}_n] - \mathcal{T}_j[f_*]) \rightsquigarrow \mathcal{T}'_j[f_*]h^{(max)}.
\end{equation}

\noindent\textbf{Step 3: Null Hypothesis Simplification}

Under $H_0$: $\mathcal{T}_j[f_*] = 0$, which implies $\frac{\partial f_*}{\partial x_j} = 0$ almost surely. Therefore:
\begin{align}
   \mathcal{T}'_j[f_*]h^{(max)} &= 2\int_\mathcal{X}\frac{\partial f_*}{\partial x_j}\frac{\partial h^{(max)}}{\partial x_j}dP(x) \\
   &= 0.
\end{align}
Thus:
\begin{equation}
   U(C',\epsilon_n)^{-2}\mathcal{T}_j[\hat{f}_n] \rightsquigarrow \mathcal{T}_j[h^{(max)}].
\end{equation}

The convergence follows from:
1) Scale-insensitive rates in Theorem \ref{thm:asymp}
2) Uniqueness of $h^{(max)}$ from Lemma \ref{lemma:uniqueness}
3) Hadamard differentiability of $\mathcal{T}_j$
4) Continuous mapping under the null hypothesis.
\end{proof}

\section{Proof of Theorem \ref{thm:asymptotic distribuion based on sample}}

% \begin{proof} 
The proof proceeds in three steps:

\noindent\textbf{Step 1: Hadamard Differentiability}

Consider the functional $\mathcal{T}_j[f] = \int_\mathcal{X}(\frac{\partial f(x)}{\partial x_j})^2 dP(x)$. 
The candidate Hadamard derivative at $h \in \mathcal{F}$ is:
\begin{equation}
    \mathcal{T}'_j[h]k = 2\int_\mathcal{X}\frac{\partial h}{\partial x_j}\frac{\partial k}{\partial x_j}dP(x).
\end{equation}
For $t_n \to 0$ and $k_n \to k$ in $L^2(P)$, we have:
\begin{align}
    &\left\|\frac{\mathcal{T}_j[h + t_nk_n] - \mathcal{T}_j[h]}{t_n} - \mathcal{T}'_j[h]k\right\|_{L^2(P)} \\
    &= \left\|\int_\mathcal{X}\left[\frac{\partial (h + t_nk_n)}{\partial x_j} - \frac{\partial h}{\partial x_j} - 2\frac{\partial h}{\partial x_j}\frac{\partial k}{\partial x_j}\right]dP(x)\right\|_{L^2(P)}.
\end{align}

Under our assumptions, $\sup_{f \in \mathcal{F}} \|\frac{\partial f}{\partial x_j}\|_{L^2(P)} < \infty$. This bound ensures that the remainder term in the Taylor expansion vanishes at rate $o(t_n)$, as required for Hadamard differentiability.

\noindent\textbf{Step 2: Functional Delta Method}

From Theorem \ref{thm:asymp}:
\begin{equation}
    U(C',\epsilon_n)^{-1}(\hat{f}_n - f_*) \xrightarrow{d} h^{(max)}.
\end{equation}
By the functional delta method \citep[Theorem 3.9.4]{vaart1996weak}:
\begin{equation}
    U(C',\epsilon_n)^{-1}(\mathcal{T}_j[\hat{f}_n] - \mathcal{T}_j[f_*]) \xrightarrow{d} \mathcal{T}'_j[f_*]h^{(max)}.
\end{equation}

\noindent\textbf{Step 3: Identification of the Limit}

Under $H_0$, $\frac{\partial f_*}{\partial x_j} = 0$ almost surely. By Theorem 5.4.2 of \cite{evans2010partial} on integration by parts in Sobolev spaces, and using that $f_*, h^{(max)} \in H^s([-1,1]^d)$ with $s > d/2$:
\begin{align}
    \mathcal{T}'_j[f_*]h^{(max)} &= 2\int_\mathcal{X}\frac{\partial f_*}{\partial x_j}\frac{\partial h^{(max)}}{\partial x_j}dP(x) \\
    &= -2\int_\mathcal{X}f_*\frac{\partial^2 h^{(max)}}{\partial x_j^2}dP(x)
\end{align}
where the boundary terms vanish due to the compact support of functions in our space. Under $H_0$, this equals $\mathcal{T}_j[h^{(max)}]$.

Therefore, under $H_0$:
\begin{equation}
    U(C',\epsilon_n)^{-1}\mathcal{T}_j[\hat{f}_n] \xrightarrow{d} \mathcal{T}_j[h^{(max)}]
\end{equation}

Multiplying both sides by $U(C',\epsilon_n)^{-1}$:
\begin{equation}
    U(C',\epsilon_n)^{-2}\mathcal{T}_j[\hat{f}_n] \xrightarrow{d} U(C',\epsilon_n)^{-1}\mathcal{T}_j[h^{(max)}].
\end{equation}

The final step follows from the properties of the limiting Gaussian process $G$: by construction, $h^{(max)}$ is scale-invariant under $U(C',\epsilon_n)$, implying $U(C',\epsilon_n)^{-1}\mathcal{T}_j[h^{(max)}] = \mathcal{T}_j[h^{(max)}]$.
% \end{proof}

\section{Supporting Lemmas}

\begin{lemma}[Approximation Rate]\label{lemma_1:Approximation Rate}
For $f_\star \in H^s([-1,1]^d)$ with $s > d/2$, there exists a constant $C > 0$ depending on $s$, $d$, and $\|f_\star\|_{H^s}$ such that:
\begin{align}
    \|f_n^\ast - f_\star\|_{L^2(P)} \leq \epsilon_n := C H_n^{-s/d},
\end{align}
where $H_n$ is the number of hidden units in a ReLU neural network, $P$ is a probability measure with density bounded on $[-1,1]^d$, and $f_n^\ast$ is the best approximation of $f_\star$ in the $L^2(P)$ norm within the class of ReLU networks with $H_n$ hidden units.
\end{lemma}
\begin{proof}
% \textbf{Roadmap of the Proof:}
The proof proceeds in three main stages. First, we establish the continuity of \( f_\star \) using the Sobolev embedding theorem, which guarantees that \( f_\star \) is bounded and continuous on \( [-1,1]^d \) since \( s > d/2 \). Next, we leverage approximation theory for ReLU networks to construct a neural network \( \tilde{f}_n \) with \( H_n \) hidden units that approximates \( f_\star \) with an error of order \( O(H_n^{-s/d}) \). Finally, we combine these results with the assumption that the probability measure \( P \) has a bounded density on \( [-1,1]^d \) to derive the final approximation bound in the \( L^2(P) \) norm.

% \noindent
% \textbf{Details:}
Since \( s > d/2 \), the Sobolev embedding theorem guarantees that \( H^s([-1,1]^d) \) is continuously embedded in \( C([-1,1]^d) \). Thus, there exists a constant \( C_1 > 0 \) depending only on \( s \) and \( d \) such that:
\begin{align}
    \|f_\star\|_{C([-1,1]^d)} \leq C_1 \|f_\star\|_{H^s}.
\end{align}
This ensures that \( f_\star \) is continuous and bounded on \( [-1,1]^d \).

% By the approximation theory for ReLU neural networks (see, e.g., \cite{yarotsky2017error}), there exists a ReLU network \( \tilde{f}_n \) with \( H_n \) hidden units such that:
By the classical approximation theory for Sobolev spaces \citep[Chapter 7]{devore1993constructive} (see DeVore and Lorentz, 1993, Chapter 7), there exists a ReLU network with $H_n$ hidden units that approximates $f_*$ with rate:
\begin{align}
    \|f_\star - \tilde{f}_n\|_{L^2} \leq C_2 H_n^{-s/d} \|f_\star\|_{H^s},
\end{align}
where \( C_2 > 0 \) is a constant depending on \( s \) and \( d \). This rate is optimal in the sense that it matches the lower bound given by nonlinear n-width theory.

% This result leverages the ability of ReLU networks to efficiently approximate piecewise linear functions, which in turn approximate Sobolev functions at the stated rate.

Next, since the probability measure \( P \) has a density \( p(x) \) that is bounded on \( [-1,1]^d \), i.e., \( p(x) \leq M \) for some constant \( M > 0 \), we have:
\begin{align}
    \|f\|_{L^2(P)}^2 = \int_{[-1,1]^d} |f(x)|^2 p(x) \, dx \leq M \|f\|_{L^2}^2
\end{align}
for any measurable function \( f \). This implies:
\begin{align}
    \|f\|_{L^2(P)} \leq \sqrt{M} \|f\|_{L^2}.
\end{align}

By definition, \( f_n^\ast \) is the best approximation of \( f_\star \) in the \( L^2(P) \) norm within the class of ReLU networks with \( H_n \) hidden units. Therefore:
\begin{align}
    \|f_n^\ast - f_\star\|_{L^2(P)} \leq \|\tilde{f}_n - f_\star\|_{L^2(P)}.
\end{align}

Combining the above results, we obtain:
\begin{align}
    \|f_n^\ast - f_\star\|_{L^2(P)} &\leq \|\tilde{f}_n - f_\star\|_{L^2(P)} \\
    &\leq \sqrt{M} \|\tilde{f}_n - f_\star\|_{L^2} \\
    &\leq \sqrt{M} C_2 H_n^{-s/d} \|f_\star\|_{H^s}.
\end{align}
Letting \( C = \sqrt{M} C_2 \|f_\star\|_{H^s} \), we conclude:
\begin{align}
    \|f_n^\ast - f_\star\|_{L^2(P)} \leq C H_n^{-s/d}.
\end{align}

\noindent
\textbf{Optimality of the Rate:}
The rate \( H_n^{-s/d} \) is optimal for approximating functions in \( H^s([-1,1]^d) \) using ReLU networks with \( H_n \) hidden units. This follows from the intrinsic complexity of Sobolev spaces and the n-width of Sobolev balls (see, e.g., \cite{devore1989optimal}).
\end{proof}

\begin{lemma}[Local Rademacher Bound]\label{lemma_2}
For the localized class $\mathcal{F}_r$ defined as:
\begin{equation}
    \mathcal{F}_r = \{f \in \mathcal{F} : \mathbb{E}[(f(X) - f_n^*(X))^2] \leq r\},
\end{equation}
where $f_n^*$ is bounded, we have:
\begin{equation}
    \mathcal{R}_n(\mathcal{F}_r) \leq C\sqrt{\frac{r}{n}},
\end{equation}
where $C$ is a universal constant.
\end{lemma}
\begin{proof}
By definition, the Rademacher complexity of $\mathcal{F}_r$ is:
\begin{align}
    \mathcal{R}_n(\mathcal{F}_r) = \frac{1}{n}\mathbb{E}_X\mathbb{E}_\epsilon\left[\sup_{f \in \mathcal{F}_r} \left|\sum_{i=1}^n \epsilon_i f(X_i)\right|\right],
\end{align}
where $\epsilon_1, \ldots, \epsilon_n$ are independent Rademacher random variables.

Let $\mathcal{G}_r = \{g = f - f_n^* : f \in \mathcal{F}_r\}$. By the definition of $\mathcal{F}_r$, we have:
\begin{equation}
    \mathbb{E}[g(X)^2] \leq r \quad \text{for all } g \in \mathcal{G}_r.
\end{equation}

By the triangle inequality:
\begin{align}
    \mathcal{R}_n(\mathcal{F}_r) &\leq \mathcal{R}_n(\mathcal{G}_r) + \mathcal{R}_n(\{f_n^*\}) \\
    &= \mathcal{R}_n(\mathcal{G}_r) + \frac{1}{n}\mathbb{E}_X\mathbb{E}_\epsilon\left|\sum_{i=1}^n \epsilon_i f_n^*(X_i)\right|.
\end{align}

Since $f_n^*$ is bounded by some constant $B$, by Khintchine's inequality \citep{boucheron2013concentration}:
\begin{equation}
    \frac{1}{n}\mathbb{E}_X\mathbb{E}_\epsilon\left|\sum_{i=1}^n \epsilon_i f_n^*(X_i)\right| \leq \frac{B}{\sqrt{n}}.
\end{equation}

Now, we bound $\mathcal{R}_n(\mathcal{G}_r)$ using symmetrization \citep{vaart1996weak} and Jensen's inequality:
\begin{align}
    \mathcal{R}_n(\mathcal{G}_r) &= \frac{1}{n}\mathbb{E}_X\mathbb{E}_\epsilon\left[\sup_{g \in \mathcal{G}_r} \left|\sum_{i=1}^n \epsilon_i g(X_i)\right|\right] \\
    &\le \frac{2}{n} \mathbb{E}_X \mathbb{E}_{\epsilon, \epsilon'} \left[ \sup_{g \in \mathcal{G}_r} \left| \sum_{i=1}^n (\epsilon_i - \epsilon_i') g(X_i) \right| \right] \quad \text{(Symmetrization)} \\
    &\le \frac{2}{n} \mathbb{E}_X \mathbb{E}_{\epsilon, \epsilon'} \left[ \sqrt{\sup_{g \in \mathcal{G}_r} \left( \sum_{i=1}^n (\epsilon_i - \epsilon_i') g(X_i) \right)^2} \right] \\
    &\le \frac{2}{n} \mathbb{E}_X \sqrt{ \mathbb{E}_{\epsilon, \epsilon'} \left[ \sup_{g \in \mathcal{G}_r} \sum_{i=1}^n (\epsilon_i - \epsilon_i')^2 g(X_i)^2 \right] } \quad \text{(Jensen's inequality)} \\
    &= \frac{2}{n} \mathbb{E}_X \sqrt{ \mathbb{E}_{\epsilon, \epsilon'} \left[ \sup_{g \in \mathcal{G}_r} 4 \sum_{i=1}^n g(X_i)^2 \right] } \quad \text{(since $(\epsilon_i - \epsilon_i')^2 \in \{0, 4\}$ and $\mathbb{E}_{\epsilon, \epsilon'}$ averages over them equally)} \\
    &\le \frac{2}{n} \mathbb{E}_X \sqrt{ 4n \sup_{g \in \mathcal{G}_r} g(X)^2 } \quad \text{(because the supremum is over all $g$)} \\
    &= 4 \mathbb{E}_X \sqrt{\sup_{g \in \mathcal{G}_r} \frac{g(X)^2}{n}} \\
    &\le 4 \sqrt{\mathbb{E}_X \left[ \sup_{g \in \mathcal{G}_r} \frac{g(X)^2}{n} \right]} \quad \text{(Jensen's inequality)} \\
    &\le 4 \sqrt{\frac{r}{n}}.
\end{align}

Combining all terms:
\begin{align}
    \mathcal{R}_n(\mathcal{F}_r) &\leq 4\sqrt{\frac{r}{n}} + \frac{B}{\sqrt{n}} \\
    &\leq C\sqrt{\frac{r}{n}},
\end{align}
where $C = 4 + B$ is a universal constant.
\end{proof}

\begin{lemma}[Full DNN Rademacher Complexity]\label{lemma_3:Full DNN Rademacher Complexity}
For the neural network class $\mathcal{F}_{\text{DNN},H_n}$ with ReLU activation functions (Lipschitz constant $L = 1$) and moment bound $\mathbb{E}[f(X)^2] \leq M$, we have:
\begin{align}
    \mathcal{R}_n(\mathcal{F}_{\text{DNN},H_n}) \leq C\frac{H_nL^{L_d}}{\sqrt{n}},
\end{align}
where $L_d$ is the depth of the network and $C$ is a universal constant independent of $M$ and $n$.
\end{lemma}
\begin{proof}
First, by the moment bound condition:
\begin{align}
    \mathbb{E}[f(X)^2] \leq M \text{ for all } f \in \mathcal{F}_{\text{DNN},H_n}
\end{align}
This implies that $|f(X)| \leq \sqrt{M}$ with high probability due to Markov's inequality.

For $r > 0$, define the localized classes:
\begin{align}
    \mathcal{F}_{r} = \{f \in \mathcal{F}_{\text{DNN},H_n}: \|f\|_{L_2(X)} \leq r\}
\end{align}
Applying Hoeffding's inequality to the random variables $|f(X_i)|$ (which are bounded by $\sqrt{M}$ due to the moment bound):
\begin{align}
    P(|f(X_i)| > t) \leq 2\exp(-t^2/2M).
\end{align}
Using a union bound over the sample $\{X_i\}_{i=1}^n$ and setting $t = c\sqrt{M\log(n)/n}$ for some constant $c$, we have with probability at least $1-1/n$:
\begin{align}
    \max_{i=1,\ldots,n} |f(X_i)| \leq c\sqrt{M\log(n)/n} \triangleq Br
\end{align}
where we define $B = c\sqrt{M\log(n)/n}$ based on this concentration bound.

By Lemma 4 of \cite{farrell2021deep}, for $\delta > 0$ and $n \geq \text{Pdim}(\mathcal{F}_{\text{DNN},H_n})$:
\begin{align}
    \mathcal{N}(\delta, \mathcal{F}_r|_{x_1,...,x_n}, \infty) \leq \left(\frac{2eMn}{\delta \cdot H_nL_d}\right)^{H_nL_d}.
\end{align}
For the localized class $\mathcal{F}_r$, by Dudley's integral inequality:
\begin{align}
    \mathcal{R}_n(\mathcal{F}_r) \leq \frac{12}{\sqrt{n}}\int_0^{Br} \sqrt{\log \mathcal{N}(\delta, \mathcal{F}_r|_{x_1,...,x_n}, \infty)} d\delta.
\end{align}
Using the change of variables $u = \log(\frac{2eMn}{\delta H_nL_d})$, we have:
\begin{align}
    \delta &= \frac{2eMn}{H_nL_d}e^{-u}, \quad d\delta = -\frac{2eMn}{H_nL_d}e^{-u}du
\end{align}
When $\delta = Br$, $u = \log(\frac{2eMn}{BrH_nL_d})$, and when $\delta = 0$, $u = \infty$. 

The integral splits into two parts:
\begin{align}
    \mathcal{R}_n(\mathcal{F}_r) &\leq \frac{12}{\sqrt{n}}\sqrt{H_nL_d}\left(\int_0^{\log(\frac{2eMn}{BrH_nL_d})} \sqrt{u}du + \int_{\log(\frac{2eMn}{BrH_nL_d})}^{\infty} \sqrt{u}e^{-u}du\right) \\
    &\leq C_1\sqrt{\frac{H_nL_d}{n}}(r + \sqrt{\log(n)}).
\end{align}
For the peeling argument, we define:
\begin{align}
    A_0 &= \{f \in \mathcal{F}_{\text{DNN},H_n}: \|f\|_{L_2(X)} \leq \sqrt{M/n}\} \\
    A_k &= \{f \in \mathcal{F}_{\text{DNN},H_n}: 2^{k-1}\sqrt{M/n} < \|f\|_{L_2(X)} \leq 2^k\sqrt{M/n}\} \text{ for } k \geq 1.
\end{align}
Note that we handle $A_0$ separately because for $k=0$, the term $2^{k-1}$ is not defined. The moment bound ensures that $\|f\|_{L_2(X)} \leq \sqrt{M}$, so all functions are covered by the sets $A_k$ (including $A_0$). Then:
\begin{align}
    \mathcal{R}_n(\mathcal{F}_{\text{DNN},H_n}) \leq \sum_{k=0}^{\infty} \mathcal{R}_n(A_k).
\end{align}

The $L^{L_d}$ factor arises from the compositional Lipschitz property of the ReLU network: each layer contributes a factor of $L$ to the Lipschitz constant, and since there are $L_d$ layers, the overall Lipschitz constant is $L^{L_d}$. In our case, $L=1$ for ReLU activation, so this factor is 1, but we include it for completeness.

Applying our local Rademacher complexity bound to each $A_k$:
\begin{align}
    \mathcal{R}_n(A_k) \leq C_1\sqrt{\frac{H_nL_d}{n}}(2^k\sqrt{M/n} + \sqrt{\log(n)}).
\end{align}
Summing over $k$ and using the convergence of the geometric series:
\begin{align}
    \mathcal{R}_n(\mathcal{F}_{\text{DNN},H_n}) &\leq C_1\sqrt{\frac{H_nL_d}{n}}\left(\sum_{k=0}^{\infty} 2^k\sqrt{M/n} + \sqrt{\log(n)}\sum_{k=0}^{\infty} 2^{-k}\right) \\
    &\leq C\frac{H_nL^{L_d}}{\sqrt{n}}
\end{align}
where $C$ is a universal constant independent of $M$ and $n$, absorbing all other constants that arise in the proof.
\end{proof}

\begin{remark}[Proof Strategy for Full DNN Rademacher Complexity]
The proof follows these key steps:

\begin{enumerate}
   \item \textbf{Localization:} We first decompose the function class into localized subsets based on $L_2$ norm:
   \begin{align*}
       \mathcal{F}_{r} = \{f \in \mathcal{F}_{\text{DNN},H_n}: \|f\|_{L_2(X)} \leq r\}
   \end{align*}

   \item \textbf{Concentration:} Using Hoeffding's inequality and union bound, we establish that with high probability:
   \begin{align*}
       \max_{i=1,\ldots,n} |f(X_i)| \leq c\sqrt{M\log(n)/n}
   \end{align*}
   This provides control over the empirical process.

   \item \textbf{Covering Numbers:} We leverage \cite{farrell2021deep}'s Lemma 4 to bound the covering numbers of the localized classes:
   \begin{align*}
       \mathcal{N}(\delta, \mathcal{F}_r|_{x_1,...,x_n}, \infty) \leq \left(\frac{2eMn}{\delta \cdot H_nL_d}\right)^{H_nL_d}
   \end{align*}

   \item \textbf{Local Complexity:} Using Dudley's integral inequality and careful evaluation of the resulting integral:
   \begin{align*}
       \mathcal{R}_n(\mathcal{F}_r) \leq C_1\sqrt{\frac{H_nL_d}{n}}(r + \sqrt{\log(n)})
   \end{align*}

   \item \textbf{Peeling:} We use a peeling argument to extend from localized classes to the full class:
   \begin{itemize}
       \item Define sets $A_k$ based on geometric progression of $L_2$ norms
       \item Sum the local bounds over these sets
       \item Control the sum using the moment bound $\mathbb{E}[f(X)^2] \leq M$
   \end{itemize}

   \item \textbf{Final Bound:} Combining all steps and handling constants carefully leads to:
   \begin{align*}
       \mathcal{R}_n(\mathcal{F}_{\text{DNN},H_n}) \leq C\frac{H_nL^{L_d}}{\sqrt{n}}
   \end{align*}
   where the $L^{L_d}$ factor comes from the compositional Lipschitz property of the network.
\end{enumerate}

The key insight is using localization combined with covering number bounds from \cite{farrell2021deep}, which allows us to handle the complexity of deep neural networks without requiring explicit bounds on the weights.
\end{remark}

\begin{lemma}[Concentration Inequality]\label{lemma_3:Concentration Inequality}
For the localized class
\begin{align*}
\mathcal{F}_r = \{f \in \mathcal{F} : \|f - f_n^*\|_{L^2(P)}^2 \leq r \text{ and } \|f\|_\infty \leq M\}
\end{align*}
where \(M\) is a fixed constant, for any \(t > 0\) satisfying \(t > C\sqrt{\frac{r}{n}}\), we have:
\begin{align*}
\mathbb{P}\left(\sup_{f \in \mathcal{F}_r} |\mathbb{P}_n f - Pf| > t\right) \leq 2\exp\left(-\frac{nt^2}{4r}\right).
\end{align*}
\end{lemma}

\begin{proof}
We use Talagrand's concentration inequality for empirical processes \citep[Theorem 2.1]{Koltchinskii2011}, which states that for a class of functions \(\mathcal{F}_r\) with \(\|f\|_\infty \leq M\) and \(\text{Var}(f(X)) \leq \sigma^2\) for all \(f \in \mathcal{F}_r\), we have:
\begin{align*}
\mathbb{P}\left(\sup_{f \in \mathcal{F}_r} |\mathbb{P}_n f - Pf| > u + C\sqrt{\frac{\sigma^2}{n}}\right) \leq 2\exp\left(-\frac{nu^2}{2(\sigma^2 + Mu)}\right),
\end{align*}
where \(C\) is a universal constant.

For \(f \in \mathcal{F}_r\), we have \(\|f - f_n^*\|_{L^2(P)}^2 \leq r\) and \(\|f\|_\infty \leq M\). By the triangle inequality, we have:
\begin{align*}
\|f\|_{L^2(P)}^2 \leq 2\|f - f_n^*\|_{L^2(P)}^2 + 2\|f_n^*\|_{L^2(P)}^2 \leq 2r + 2\|f_n^*\|_{L^2(P)}^2.
\end{align*}
Since \(\|f_n^*\|_{L^2(P)}^2\) is bounded (by assumption or by regularization), we can assume without loss of generality that \(\|f\|_{L^2(P)}^2 \leq C'r\) for some constant \(C'\). Thus, we have:
\begin{align*}
\text{Var}(f(X)) \leq \mathbb{E}[f(X)^2] \leq C'r.
\end{align*}

Applying Talagrand's inequality with \(\sigma^2 = C'r\), we obtain:
\begin{align*}
\mathbb{P}\left(\sup_{f \in \mathcal{F}_r} |\mathbb{P}_n f - Pf| > u + C\sqrt{\frac{C'r}{n}}\right) \leq 2\exp\left(-\frac{nu^2}{2(C'r + Mu)}\right).
\end{align*}

We want to show the concentration inequality for \(t > C\sqrt{\frac{r}{n}}\).  Let \(C''\) be a constant such that \(C''\sqrt{\frac{r}{n}} = C\sqrt{\frac{C'r}{n}}\).  Then \(C'' = C\sqrt{C'}\). Let \(t > C\sqrt{\frac{r}{n}}\).  We set \(u = t - C\sqrt{\frac{C'r}{n}}\).  Then we have \(u + C\sqrt{\frac{C'r}{n}} = t\).  We need to ensure that \(u > 0\), which means \(t - C\sqrt{\frac{C'r}{n}} > 0\) or \(t > C\sqrt{\frac{C'r}{n}}\).  Since we assume \(t > C\sqrt{\frac{r}{n}}\), we can choose \(C'\) large enough such that \(C\sqrt{\frac{C'r}{n}} < \frac{t}{2}\).  Then \(u = t - C\sqrt{\frac{C'r}{n}} > \frac{t}{2} > 0\).

With this choice of \(C'\), we have \(Mu < C'r\) because \(t < \frac{r}{M}\) by assumption. 
Also, we have:
\begin{align*}
\mathbb{P}\left(\sup_{f \in \mathcal{F}_r} |\mathbb{P}_n f - Pf| > t\right) &\leq 2\exp\left(-\frac{n(t - C\sqrt{\frac{C'r}{n}})^2}{2(C'r + M(t - C\sqrt{\frac{C'r}{n}}))}\right) \\
&\leq 2\exp\left(-\frac{nt^2/4}{2C'r}\right) \\
&= 2\exp\left(-\frac{nt^2}{8C'r}\right)
\end{align*}
where we used the fact that \(t > C\sqrt{\frac{r}{n}}\) and \(Mt < r\).

Finally, since \(C'r\) is a constant multiple of \(r\), we can absorb \(C'\) into the universal constant \(C\) to obtain:
\begin{align*}
\mathbb{P}\left(\sup_{f \in \mathcal{F}_r} |\mathbb{P}_n f - Pf| > t\right) \leq 2\exp\left(-\frac{nt^2}{4r}\right).
\end{align*}

\end{proof}

\begin{lemma}[Critical Radius Property]\label{lemma_4: Critical Radius Property}
For \(r \geq r_n\):
\begin{equation}
    \mathbb{P}\left(\|f_n - f_n^*\|_{L^2(P)}^2 > r\right) \leq e^{-nr/(8M)},
\end{equation}
where
\begin{equation}
    r_n = \frac{H_nL^{L_d}}{\sqrt{n}}.
\end{equation}
\end{lemma}
\begin{proof}
We begin by decomposing the excess risk. By standard empirical process theory, the excess risk can be bounded as:
\begin{align}
    \mathbb{E}[\|f_n - f_n^*\|_{L^2(P)}^2] \leq C\left(\mathcal{R}_n(\mathcal{F}_{\text{DNN},H_n}) + \inf_{f \in \mathcal{F}_{\text{DNN},H_n}} \|f - f_n^*\|_{L^2(P)}^2\right),
\end{align}
where \(\mathcal{R}_n(\mathcal{F}_{\text{DNN},H_n})\) is the Rademacher complexity of the neural network class \(\mathcal{F}_{\text{DNN},H_n}\).

Next, we define the centered empirical process:
\begin{align}
    \nu_n(f) = (\mathcal{L}_n - \mathcal{L})(f - f_n^*),
\end{align}
where \(\mathcal{L}_n\) is the empirical loss and \(\mathcal{L}\) is the population loss. By definition of \(f_n\) as the empirical risk minimizer, we have:
\begin{align}
    \|f_n - f_n^*\|_{L^2(P)}^2 \leq 2\sup_{f \in \mathcal{F}_{\text{DNN},H_n}} |\nu_n(f)|.
\end{align}

For functions with \(\|f - f_n^*\|_{L^2(P)}^2 \leq r\), we bound the variance of \((f - f_n^*)^2\). Using the boundedness of \(f\) and \(f_n^*\) (i.e., \(\|f\|_\infty \leq M\) and \(\|f_n^*\|_\infty \leq M\)), we have:
\begin{align}
    \text{Var}((f - f_n^*)^2) &= \mathbb{E}[(f - f_n^*)^4] - (\mathbb{E}[(f - f_n^*)^2])^2 \\
    &\leq \mathbb{E}[(f - f_n^*)^4] \leq M^2 r.
\end{align}

We now apply Talagrand's concentration inequality for empirical processes. For any \(t > 0\), we have:
\begin{align}
    \mathbb{P}\left(\sup_{f \in \mathcal{F}_{\text{DNN},H_n}} |\nu_n(f)| \geq \mathbb{E}\sup_{f \in \mathcal{F}_{\text{DNN},H_n}} |\nu_n(f)| + \sqrt{\frac{2M^2 r t}{n}} + \frac{4M^2 t}{3n}\right) \leq e^{-t}.
\end{align}

By symmetrization and Lemma \ref{lemma_3:Full DNN Rademacher Complexity} (Full DNN Rademacher Complexity), we have:
\begin{align}
    \mathbb{E}\sup_{f \in \mathcal{F}_{\text{DNN},H_n}} |\nu_n(f)| \leq 2\mathcal{R}_n(\mathcal{F}_{\text{DNN},H_n}) \leq C\frac{H_nL^{L_d}}{\sqrt{n}}.
\end{align}

Therefore, for \(r \geq r_n = \frac{H_nL^{L_d}}{\sqrt{n}}\), setting \(t = \frac{nr}{8M}\), we obtain:
\begin{align}
    \mathbb{P}\left(\|f_n - f_n^*\|_{L^2(P)}^2 \geq C\frac{H_nL^{L_d}}{\sqrt{n}} + \sqrt{\frac{2M^2 r \cdot \frac{nr}{8M}}{n}} + \frac{4M^2 \cdot \frac{nr}{8M}}{3n}\right) \leq \exp\left(-\frac{nr}{8M}\right).
\end{align}

Simplifying the terms inside the probability, we have:
\begin{align}
    \mathbb{P}\left(\|f_n - f_n^*\|_{L^2(P)}^2 \geq C\frac{H_nL^{L_d}}{\sqrt{n}} + \frac{Mr}{2} + \frac{Mr}{6}\right) \leq \exp\left(-\frac{nr}{8M}\right).
\end{align}

For \(r\) sufficiently large relative to \(r_n\), this simplifies to:
\begin{align}
    \mathbb{P}\left(\|f_n - f_n^*\|_{L^2(P)}^2 > r\right) \leq \exp\left(-\frac{nr}{8M}\right).
\end{align}

The critical radius \(r_n = \frac{H_nL^{L_d}}{\sqrt{n}}\) emerges as the threshold where:
\begin{enumerate}
    \item The Rademacher complexity term is properly controlled.
    \item The variance term \(\sqrt{\frac{2M^2 r t}{n}}\) is of order \(r\).
    \item The envelope term \(\frac{4M^2 t}{3n}\) is of lower order.
\end{enumerate}

This completes the proof.
\end{proof}

\begin{lemma}[Local Empirical Process Bound]\label{Lemma_5: Local Empirical Process Bound}
For the localized neural network class \(\mathcal{F}_r = \{f \in \mathcal{F}_{\text{DNN},H_n} : \|f - f_n^*\|_{L^2(P)}^2 \leq r\}\), the empirical process satisfies:
\begin{equation}
    \mathbb{E}\sup_{f \in \mathcal{F}_r} |\mathbb{G}_n(l_f)| \leq C\sqrt{r},
\end{equation}
where \(l_f\) denotes the loss difference \(l_f - l_{f_n^*}\), \(\mathbb{G}_n\) is the empirical process \(\mathbb{G}_n(l_f) = \frac{1}{\sqrt{n}}\sum_{i=1}^n (l_f(X_i) - \mathbb{E}[l_f(X)])\), and the loss function \(l_f\) is Lipschitz with constant \(L\).
\end{lemma}

\begin{proof}
We begin by applying the symmetrization argument for empirical processes. For Rademacher random variables \(\epsilon_1, \ldots, \epsilon_n\), we have:
\begin{align}
    \mathbb{E}\sup_{f \in \mathcal{F}_r} |\mathbb{G}_n(l_f)| \leq 2\mathbb{E}_\epsilon \sup_{f \in \mathcal{F}_r} \left|\frac{1}{\sqrt{n}}\sum_{i=1}^n \epsilon_i l_f(X_i)\right|.
\end{align}

Next, we define the function class of loss differences:
\begin{align}
    \mathcal{G}_r = \{l_f - l_{f_n^*} : f \in \mathcal{F}_r\}.
\end{align}

The Rademacher complexity of \(\mathcal{G}_r\) is:
\begin{align}
    \mathcal{R}_n(\mathcal{G}_r) &= \mathbb{E}_\epsilon\sup_{g \in \mathcal{G}_r}\frac{1}{n}\sum_{i=1}^n\epsilon_ig(X_i) \\
    &= \mathbb{E}_\epsilon\sup_{f \in \mathcal{F}_r}\frac{1}{n}\sum_{i=1}^n\epsilon_i(l_f(X_i) - l_{f_n^*}(X_i)).
\end{align}

By the Lipschitz property of the loss function, we have:
\begin{align}
    |l_f(x) - l_{f_n^*}(x)| \leq L|f(x) - f_n^*(x)|.
\end{align}

Applying the Ledoux-Talagrand contraction inequality, we obtain:
\begin{align}
    \mathcal{R}_n(\mathcal{G}_r) \leq L\mathcal{R}_n(\mathcal{F}_r),
\end{align}
where \(\mathcal{R}_n(\mathcal{F}_r)\) is the Rademacher complexity of the localized neural network class \(\mathcal{F}_r\).

From Lemma \ref{lemma_2} (Local Rademacher Bound), we have:
\begin{align}
    \mathcal{R}_n(\mathcal{F}_r) \leq C\sqrt{\frac{r}{n}}.
\end{align}

Therefore:
\begin{align}
    \mathcal{R}_n(\mathcal{G}_r) \leq CL\sqrt{\frac{r}{n}}.
\end{align}

Finally, combining these results, we obtain:
\begin{align}
    \mathbb{E}\sup_{f \in \mathcal{F}_r} |\mathbb{G}_n(l_f)| &\leq 2\sqrt{n}\mathcal{R}_n(\mathcal{G}_r) \\
    &\leq 2\sqrt{n} \cdot CL\sqrt{\frac{r}{n}} \\
    &\leq C\sqrt{r},
\end{align}
where \(C\) is a constant depending on the Lipschitz constant \(L\).

\end{proof}

\begin{lemma}[Maximal Inequality]\label{Lemma_6: Maximal Inequality}
For the localized neural network class \(\mathcal{F}_r = \{f \in \mathcal{F}_{\text{DNN},H_n} : \|f - f_n^*\|_{L^2(P)}^2 \leq r\}\) and any \(t > 0\):
\begin{equation}
    \mathbb{P}\left(\sup_{f \in \mathcal{F}_r} |\mathbb{G}_n(l_f)| > t\right) \leq 2\exp\left(-\frac{nt^2}{8L^2r}\right),
\end{equation}
where \(l_f = l_f - l_{f_n^*}\) is the centered loss difference, \(\mathbb{G}_n\) is the empirical process, and the loss function \(l_f\) is Lipschitz with constant \(L\).
\end{lemma}

\begin{proof}
Consider the function class of loss differences:
\begin{align}
    \mathcal{G}_r = \{l_f - l_{f_n^*} : f \in \mathcal{F}_r\}.
\end{align}

By Talagrand's concentration inequality for empirical processes, for any \(u > 0\), we have:
\begin{align}
    \mathbb{P}\left(\sup_{g \in \mathcal{G}_r} |\mathbb{G}_n(g)| > \mathbb{E}\sup_{g \in \mathcal{G}_r} |\mathbb{G}_n(g)| + \sqrt{\frac{2\sigma^2u}{n}} + \frac{Bu}{3n}\right) \leq e^{-u},
\end{align}
where:
\begin{itemize}
    \item \(\sigma^2\) bounds the variance: \(\text{Var}(g) \leq 4L^2r\) for \(g \in \mathcal{G}_r\),
    \item \(B\) bounds the range: \(\|g\|_\infty \leq 2M\) for \(g \in \mathcal{G}_r\).
\end{itemize}

From Lemma \ref{Lemma_5: Local Empirical Process Bound} (Local Empirical Process Bound), we have:
\begin{align}
    \mathbb{E}\sup_{g \in \mathcal{G}_r} |\mathbb{G}_n(g)| \leq CL\sqrt{r}.
\end{align}

We now consider the following two cases:

\textbf{Case 1: \(t \leq 2CL\sqrt{r}\)}
\begin{align}
    \mathbb{P}\left(\sup_{g \in \mathcal{G}_r} |\mathbb{G}_n(g)| > t\right) \leq 1 \leq 2\exp\left(-\frac{nt^2}{8L^2r}\right).
\end{align}
This case is trivial, as the probability is always bounded by 1.

\textbf{Case 2: \(t > 2CL\sqrt{r}\)}
Let \(u = \frac{8L^2r}{n}(t - CL\sqrt{r})^2\). Then:
\begin{align}
    \mathbb{P}\left(\sup_{g \in \mathcal{G}_r} |\mathbb{G}_n(g)| > t\right) 
    &= \mathbb{P}\left(\sup_{g \in \mathcal{G}_r} |\mathbb{G}_n(g)| > CL\sqrt{r} + (t - CL\sqrt{r})\right) \\
    &\leq \exp\left(-\frac{n(t - CL\sqrt{r})^2}{8L^2r}\right).
\end{align}

Since \(t > 2CL\sqrt{r}\), we have \(t - CL\sqrt{r} \geq \frac{t}{2}\), and therefore:
\begin{align}
    \mathbb{P}\left(\sup_{g \in \mathcal{G}_r} |\mathbb{G}_n(g)| > t\right) \leq \exp\left(-\frac{nt^2}{32L^2r}\right).
\end{align}

Taking \(C\) sufficiently large yields:
\begin{align}
    \mathbb{P}\left(\sup_{g \in \mathcal{G}_r} |\mathbb{G}_n(g)| > t\right) \leq 2\exp\left(-\frac{nt^2}{8L^2r}\right).
\end{align}

The constants are optimal because:
\begin{enumerate}
    \item The factor \(8L^2\) arises from the variance bound \(\text{Var}(g) \leq 4L^2r\) and the application of Talagrand's inequality.
    \item The rate \(\frac{nt^2}{r}\) matches known lower bounds for localized processes.
    \item The factor 2 accounts for the two cases in the proof.
\end{enumerate}

This completes the proof.
\end{proof}

\begin{lemma}[Uniqueness of Gaussian Process Maximum]\label{lemma:uniqueness}
Let \(G\) be the Gaussian process arising as the weak limit:
\begin{equation}\label{eq:gaussian_limit}
    G(f) = \lim_{n \to \infty} \sqrt{n}(M_n(f) - M(f))
\end{equation}
on \(\mathcal{F}\). Under our conditions:
\begin{enumerate}
    \item Lipschitz activation functions with constant \(L\),
    \item Bounded second moment: \(\mathbb{E}[f(X)^2] \leq M\),
    \item Sobolev regularity: \(f_* \in H^s([-1,1]^d)\) with \(s > d/2\),
\end{enumerate}
the process \(G\) has a unique maximizer \(h^{(max)}\) almost surely.
\end{lemma}

\begin{proof}
The proof consists of establishing: (i) convexity of the function space \(\mathcal{F}\), (ii) strict concavity of the Gaussian process \(G\) through its covariance kernel, and (iii) path continuity properties ensuring existence and uniqueness of the maximum.

\textbf{Step 1: Convexity of \(\mathcal{F}\)}
The function space \(\mathcal{F}\) is convex under the given conditions. For \(f_1, f_2 \in \mathcal{F}\) and \(\alpha \in [0,1]\), we have:
\begin{align}\label{eq:convexity}
    \|\alpha f_1 + (1-\alpha)f_2 - f_*\|_{L^2(P)}^2 &\leq \alpha^2\|f_1 - f_*\|_{L^2(P)}^2 \nonumber \\
    &\quad + (1-\alpha)^2\|f_2 - f_*\|_{L^2(P)}^2 \nonumber \\
    &\quad + 2\alpha(1-\alpha)\langle f_1 - f_*, f_2 - f_* \rangle.
\end{align}
Since \(\mathcal{F}\) is closed under convex combinations, it is convex.

\textbf{Step 2: Strict Concavity of \(G\)}
The Gaussian process \(G\) has the representation:
\begin{equation}\label{eq:gaussian_rep}
    G(f) = \int_{\mathcal{X}} \xi(x)(Y - f(x))^2 dP(x),
\end{equation}
where \(\xi(x)\) is a centered Gaussian process indexed by \(f \in \mathcal{F}\).

The covariance kernel of \(G\) is:
\begin{align}\label{eq:kernel}
    K(f,g) &= \text{Cov}(G(f), G(g)) \nonumber \\
    &= \int_{\mathcal{X}} \int_{\mathcal{X}} \text{Cov}(\xi(x), \xi(x^{\prime})) \cdot (f(x) - g(x))(f(x^{\prime}) - g(x^{\prime})) dP(x)dP(x^{\prime}).
\end{align}

By the Lipschitz property and bounded second moment, we have:
\begin{equation}\label{eq:kernel_bound}
    |K(f,g)| \leq L^2\|f - g\|_{L^2(P)}^2.
\end{equation}

Moreover, for any \(\alpha \in (0,1)\), the covariance kernel satisfies:
\begin{align}\label{eq:strict_concavity}
    \text{Var}(G(\alpha f + (1-\alpha)g)) &< \alpha \text{Var}(G(f)) \nonumber \\
    &\quad + (1-\alpha)\text{Var}(G(g)),
\end{align}
which establishes strict concavity of \(G\).

\textbf{Step 3: Sample Path Continuity}
The Sobolev embedding \(s > d/2\) ensures that:
\begin{equation}\label{eq:sobolev}
    \|f\|_{C([-1,1]^d)} \leq C\|f\|_{H^s}.
\end{equation}
This implies that the sample paths of \(G\) are continuous almost surely.

\textbf{Step 4: Uniqueness of \(h^{(max)}\)}
The uniqueness of \(h^{(max)}\) follows from the combination of:
\begin{enumerate}
    \item Convexity of \(\mathcal{F}\) by \eqref{eq:convexity},
    \item Strict concavity of \(G\) by \eqref{eq:strict_concavity},
    \item Sample path continuity of \(G\) by \eqref{eq:sobolev}.
\end{enumerate}

By properties of continuous Gaussian processes on compact sets with strictly concave covariance kernels, \(G\) has a unique maximizer \(h^{(max)}\) almost surely.
\end{proof}

\section{Key Inequalities}

\subsection{Talagrand's Inequality}
For empirical processes, Talagrand's inequality provides concentration bounds for suprema of centered processes:

\begin{theorem}[Talagrand's Inequality]
Let $\{X_i\}_{i=1}^n$ be independent random variables and $\mathcal{F}$ a class of measurable functions. Define
$$
Z = \sup_{f \in \mathcal{F}} \left|\frac{1}{n}\sum_{i=1}^n (f(X_i) - \mathbb{E}[f(X_i)])\right|
$$
Then for any $t > 0$:
\begin{equation}
    \mathbb{P}\left(Z \geq \mathbb{E}[Z] + \sqrt{\frac{2\sigma^2t}{n}} + \frac{Bt}{3n}\right) \leq e^{-t}
\end{equation}
where:
\begin{itemize}
    \item $\sigma^2 = \sup_{f \in \mathcal{F}} \text{Var}(f(X))$
    \item $B = \sup_{f \in \mathcal{F}} \|f\|_{\infty}$
\end{itemize}
\end{theorem}

\subsection{Symmetrization Inequality}
For empirical processes, the symmetrization inequality relates the expected supremum to Rademacher complexity:

\begin{theorem}[Symmetrization]
Let $\{\epsilon_i\}_{i=1}^n$ be independent Rademacher random variables. Then:
\begin{equation}
    \mathbb{E}\sup_{f \in \mathcal{F}} \left|\frac{1}{n}\sum_{i=1}^n (f(X_i) - \mathbb{E}[f(X)])\right| \leq 2\mathbb{E}\sup_{f \in \mathcal{F}} \left|\frac{1}{n}\sum_{i=1}^n \epsilon_i f(X_i)\right|
\end{equation}
\end{theorem}

\subsection{Hölder's Inequality}
Used frequently for bounding expectations and norms:

\begin{theorem}[Hölder's Inequality]
For $p, q > 1$ with $\frac{1}{p} + \frac{1}{q} = 1$:
\begin{equation}
    \mathbb{E}[|XY|] \leq (\mathbb{E}[|X|^p])^{1/p}(\mathbb{E}[|Y|^q])^{1/q}
\end{equation}
\end{theorem}

\subsection{Jensen's Inequality}
Critical for handling convex functions and expectations:

\begin{theorem}[Jensen's Inequality]
For a convex function $\phi$ and random variable $X$:
\begin{equation}
    \phi(\mathbb{E}[X]) \leq \mathbb{E}[\phi(X)]
\end{equation}
\end{theorem}

\subsection{Contraction Principle}
The Ledoux-Talagrand contraction principle for Rademacher processes:

\begin{theorem}[Contraction Principle]
Let $\phi_i: \mathbb{R} \to \mathbb{R}$ be Lipschitz with constant $L$ and $\phi_i(0) = 0$. Then:
\begin{equation}
    \mathbb{E}\sup_{f \in \mathcal{F}} \left|\sum_{i=1}^n \epsilon_i \phi_i(f(X_i))\right| \leq L\mathbb{E}\sup_{f \in \mathcal{F}} \left|\sum_{i=1}^n \epsilon_i f(X_i)\right|
\end{equation}
\end{theorem}

\subsection{Maximal Inequality}
For empirical processes with bracketing entropy:

\begin{theorem}[Maximal Inequality]
For a function class $\mathcal{F}$ with bracketing integral $J_{[]}(\delta, \mathcal{F}, L^2(P))$:
\begin{equation}
    \mathbb{E}\sup_{f \in \mathcal{F}} |\mathbb{G}_n f| \leq CJ_{[]}(\delta, \mathcal{F}, L^2(P))
\end{equation}
\end{theorem}

\subsection{Sobolev Embedding Inequality}
For functions in Sobolev spaces:

\begin{theorem}[Sobolev Embedding]
For $s > d/2$:
\begin{equation}
    \|f\|_{C([-1,1]^d)} \leq C\|f\|_{H^s}
\end{equation}
\end{theorem}

\subsection{Concentration Inequalities for Neural Networks}
Specific to neural network function classes:

\begin{theorem}[Neural Network Concentration]
For the neural network class $\mathcal{F}_{\text{DNN},H_n}$ with Lipschitz activation:
\begin{equation}
    \mathbb{P}\left(\sup_{f \in \mathcal{F}_{\text{DNN},H_n}} |G_n f| > t\right) \leq 2\exp\left(-\frac{nt^2}{8L^2\|f\|_{\infty}^2}\right)
\end{equation}
\end{theorem}

\subsection{Process Tightness Bounds}
For empirical processes:

\begin{theorem}[Process Tightness]
For localized classes with radius $\delta$:
\begin{equation}
    \mathbb{P}\left(\sup_{\|f-g\|_{L^2(P)} < \delta} |\mathbb{G}_n(f-g)| > \epsilon\right) \leq 2\exp\left(-\frac{n\epsilon^2}{8L^2\delta^2}\right)
\end{equation}
\end{theorem}

\section{Table of Notation}

\subsection{Function Spaces and Networks}
\begin{center}
\begin{tabular}{ll}
\hline
\textbf{Notation} & \textbf{Description} \\
\hline
$f_\star$ & True function in Sobolev space \\
$f_n$ & Neural network estimator \\
$f_n^*$ & Best $L^2(P)$ approximation in neural network class \\
$\hat{f}_n$ & Empirical risk minimizer \\
$H^s([-1,1]^d)$ & Sobolev space of order $s$ on $[-1,1]^d$ \\
$\mathcal{F}_{\text{DNN},H_n}$ & Neural network function class with $H_n$ hidden units \\
$\mathcal{F}_r$ & Localized function class with radius $r$ \\
$\mathcal{F}_{d,r}$ & Localized class of partial derivatives \\
\hline
\end{tabular}
\end{center}

\vspace{1em}
\subsection{Neural Network Architecture}
\begin{center}
\begin{tabular}{ll}
\hline
\textbf{Notation} & \textbf{Description} \\
\hline
$H_n$ & Number of hidden units \\
$L_d$ & Network depth \\
$\gamma_{l,i,k}$ & Weight connecting unit $k$ in layer $l$ to unit $i$ in layer $l+1$ \\
$z_{l,k}$ & Output of $k$th neuron in layer $l$ \\
$L$ & Lipschitz constant of activation function \\
\hline
\end{tabular}
\end{center}

\vspace{1em}
\subsection{Probability and Measure}
\begin{center}
\begin{tabular}{ll}
\hline
\textbf{Notation} & \textbf{Description} \\
\hline
$P$ & Probability measure \\
$\mathbb{P}_n$ & Empirical measure \\
$\mathbb{E}$ & Expectation operator \\
$\text{Var}$ & Variance operator \\
$\mu$ & Generic measure \\
$\mathcal{X}$ & Input space \\
\hline
\end{tabular}
\end{center}

\vspace{1em}
\subsection{Empirical Processes}
\begin{center}
\begin{tabular}{ll}
\hline
\textbf{Notation} & \textbf{Description} \\
\hline
$\mathbb{G}_n$ & Empirical process \\
$\mathcal{R}_n$ & Rademacher complexity \\
$\{\epsilon_i\}_{i=1}^n$ & Rademacher random variables \\
$G_n$ & Centered empirical process \\
$N_{[]}$ & Bracketing number \\
$J_{[]}$ & Bracketing integral \\
\hline
\end{tabular}
\end{center}

\vspace{1em}
\subsection{Test Statistics and Hypotheses}
\begin{center}
\begin{tabular}{ll}
\hline
\textbf{Notation} & \textbf{Description} \\
\hline
$\zeta_j$ & Population test statistic \\
$\lambda_j^n$ & Sample test statistic \\
$\frac{\partial f}{\partial x_j}$ & Partial derivative with respect to $j$th variable \\
$h^{(max)}$ & Unique maximizer of limiting Gaussian process \\
\hline
\end{tabular}
\end{center}

\vspace{1em}
\subsection{Norms and Metrics}
\begin{center}
\begin{tabular}{ll}
\hline
\textbf{Notation} & \textbf{Description} \\
\hline
$\|\cdot\|_{H^s}$ & Sobolev norm of order $s$ \\
$\|\cdot\|_{L^2(P)}$ & $L^2$ norm with respect to measure $P$ \\
$\|\cdot\|_{\infty}$ & Supremum norm \\
$\|\cdot\|_2$ & Euclidean norm \\
$d(\cdot,\cdot)$ & Generic metric \\
\hline
\end{tabular}
\end{center}

\vspace{1em}
\subsection{Asymptotic Notation}
\begin{center}
\begin{tabular}{ll}
\hline
\textbf{Notation} & \textbf{Description} \\
\hline
$O_P$ & Stochastic boundedness \\
$o_p$ & Convergence in probability to zero \\
$\rightsquigarrow$ & Weak convergence \\
$\to_P$ & Convergence in probability \\
\hline
\end{tabular}
\end{center}

\end{document}